\pdfoutput=1
\documentclass[11pt]{article}

\usepackage[preprint]{acl}

\usepackage{times}
\usepackage{latexsym}

\usepackage[T1]{fontenc}
\usepackage[utf8]{inputenc}

\usepackage{microtype}

\usepackage{inconsolata}

\usepackage{graphicx}

\usepackage{amsmath}
\usepackage{amsfonts}
\usepackage{amssymb}
\usepackage{booktabs}
\usepackage{subcaption}
\makeatletter
\@ifundefined{tbl_save_outer_table_cols:}
  {\IfFileExists{array-2023-11-01.sty}{\usepackage{array}[=v2.5]}{\usepackage{array}}}
  {\usepackage{array}}
\makeatother
\usepackage{url}
\usepackage{pifont, xcolor}
\usepackage{colortbl}

\newcommand{\ds}{DocSplit}
\newcommand{\cbench}{ConfBench}
\newcommand{\cmark}{\textcolor{green}{\ding{51}}}
\newcommand{\xmark}{\textcolor{red}{\ding{55}}}

\definecolor{revision-blue}{RGB}{0,0,200} 
\newif\ifshowrevisions
\showrevisionsfalse 
\ifshowrevisions
  \DeclareRobustCommand{\revision}[1]{{\color{revision-blue}#1}}
\else
  \DeclareRobustCommand{\revision}[1]{#1}
\fi

\newif\ifanonbuild
\makeatletter
\ifacl@anonymize\anonbuildtrue\else\anonbuildfalse\fi
\makeatother
\newcommand{\dataseturl}{https://huggingface.co/datasets/amazon/ConfBench}
\ifanonbuild
  \newcommand{\datasetfootnote}{}                 
  \newcommand{\dataseturlinline}{}                 
  \newcommand{\releaseverb}{plan to release}       
  \newcommand{\releasedstate}{to be released}      
  \newcommand{\releasecontains}{will contain}      
  \newcommand{\releaseattributes}{will attribute}
  \newcommand{\releaserecords}{will record}
  \newcommand{\releasepublish}{will publish}
\else
  \newcommand{\datasetfootnote}{\footnote{\url{\dataseturl}}}
  \newcommand{\dataseturlinline}{\ \url{\dataseturl}}
  \newcommand{\releaseverb}{release}
  \newcommand{\releasedstate}{released}
  \newcommand{\releasecontains}{contains}
  \newcommand{\releaseattributes}{attributes}
  \newcommand{\releaserecords}{records}
  \newcommand{\releasepublish}{publish}
\fi

\title{\textit{Can You Trust the Confidence?} {\cbench} for \\
Vision-Language Models on Document Extraction}

\author{\textbf{Priyashree Roy, Sujitha Martin, Mohammad Rostami, Spencer Romo, Renhao Xue,} \\
  \textbf{Bob Strahan, Diego A. Socolinsky, Boyi Xie, Md Mofijul Islam} \\
  \textbf{Amazon Web Services}}

\begin{document}
\maketitle

\begin{abstract}

Intelligent document processing (IDP) with vision-language models (VLMs) hinges on confidence scores trustworthy enough to route extractions between automation and human review. Existing document benchmarks are dominated by clean, high-quality samples, leaving low accuracy regions too sparse for calibration assessment. We introduce \textbf{\cbench}, the first calibration-specific benchmark for key information extraction (KIE), built by applying 20 controlled degradation pipelines to a diverse document set, yielding 1,346 variants and 70K+ entity-level evaluations spanning the full accuracy spectrum. We evaluate four proprietary and three open-weight VLMs under verbalized and log-probability confidence estimation methods across three input modalities, and find: 
(i) OCR+Image modality results in more accurate confidence estimates; \revision{(ii) model capability is the dominant factor: within the Claude family confidence quality scales monotonically with capability, while across families parameter count is a poor predictor; (iii) calibration quality varies widely across models, from near-perfect to severely overconfident, and per-model post-hoc correction rescales these absolute confidence values for threshold-based routing without altering ranking-based operational metrics; and (iv) log-probability with first-token aggregation consistently outperforms mean-token and margin aggregations.} We also introduce \textit{ECARB}, a review-budget metric translating discriminative gains into operational savings. \revision{We \releaseverb{} \cbench{} publicly\datasetfootnote} to enable systematic study of confidence estimators and calibration methods for trustworthy IDP application deployment.

\end{abstract}
\section{Introduction}
\label{sec:intro}

 Intelligent document processing (IDP) systems powered by multimodal foundation models extract structured information from invoices, contracts, and other enterprise documents at scale~\cite{mandvikar2023augmenting}. Their business value depends on trust: organizations need reliable confidence estimates 
 to route high-certainty extractions to automation while flagging uncertain extractions for human review. Without reliable and calibrated confidence scores, organizations either need to review most documents manually that limits automation, or accept unknown error rates that risk compliance violations and financial losses~\cite{natarajan2025human,guo2017calibration,kim2022ocr}.

   While confidence calibration is well studied in image classification and NLP~\cite{guo2017calibration,joy2023sample,balanya2024adaptive}, IDP lacks both suitable benchmarks and systematic evaluation. Existing document benchmarks focus on clean, high-quality documents that yield predominantly high-accuracy predictions, leaving low- and mid-accuracy regions too sparse for calibration assessment~\cite{townsend2024realkie,ouyang2025omnidocbench}. Real world documents, however, exhibit substantial quality variation from scanning artifacts and physical degradation that directly impacts both extraction accuracy and confidence estimation. Although two families of confidence estimation have emerged for LLMs, i.e., verbalized confidence, where models self-report numeric
   scores~\cite{tian2023just,xiong2024can}, and token-level log-probability methods~\cite{ma2025estimating}, their comparative effectiveness for structured document extraction remains unexplored. No
   systematic benchmark exists to evaluate confidence calibration for key information extraction in IDP.

   \begin{table*}[!t]
   \centering
   \small
   \begin{tabular}{@{}p{2cm}cp{9.5cm}@{}}
   \toprule
   \textbf{Tier} & \textbf{Count} & \textbf{Examples} \\
   \midrule
   \cellcolor{red!20}\textbf{High}\newline($>$10\%) & 2 & \texttt{Custom22} (21\%): dithering + dot-matrix degrade text regions. \newline \texttt{Archetype9} (14\%): letterpress, fax artifacts, and high
   geometric distortion. \\
   \addlinespace
   \cellcolor{yellow!20}\textbf{Moderate}\newline(5--10\%) & 11 & Largest tier by design, covering common real-world degradation. \newline \texttt{Custom12} (9\%): dirty drum and roller streaks. \newline
   \texttt{Custom18} (8\%): rotated scan with page border. \\
   \addlinespace
   \cellcolor{gray!20}\textbf{Mild}\newline($<$5\%) & 7 & \texttt{Custom15} (4\%): moir\'{e} pattern on colored paper. \newline \texttt{Custom17} (2\%): JPEG compression with subtle noise. \\
   \bottomrule
   \end{tabular}
   \caption{Degradation tiers with representative augmentation pipelines. Accuracy drop is averaged across nine runs (only pipelines with $N > 1{,}000$ entities per run included).
   See \textbf{Appendix~\ref{sec:appendix_augmentation_examples}} for document samples and \textbf{Appendix~\ref{sec:appendix_accuracy_drop}} for accuracy impact analysis details.}
   \label{tab:degradation_tiers}
   \end{table*}

   We address these limitations through three contributions. First, we introduce \textbf{\cbench}, the first benchmark for evaluating confidence
   calibration in IDP. We apply 20 controlled degradation pipelines to a verified document subset, generating 1,346 document variants spanning the full accuracy spectrum with over 70K entity-level evaluations
   after quality filtration. Second, we evaluate seven foundation models, four proprietary and three open-weight, under verbalized confidence (1S-TopK) across three input modalities, establishing the first empirical baselines for  VLM confidence calibration on KIE. Third, for the two open-weight models exposing token-level probabilities, we compare three log-probability aggregation methods (first-token, mean-token, and margin) against verbalized confidence, revealing fundamental differences in discriminative ability between the two confidence families. We also introduce \textbf{ECARB}, a review-budget metric that translates discriminative ability into operational savings for human-in-the-loop deployments.\newline
\revision{Our evaluation yields four findings: (1) OCR+Image is uniformly the strongest input modality for both extraction accuracy and confidence quality, and the gap to image-only is largest for smaller, lower-capability models; (2) for frontier proprietary models, where token-level probabilities are not exposed, verbalized confidence tracks extraction accuracy closely and Claude Opus is nearly well-calibrated without any post-hoc adjustment, making verbalized scores a viable production signal for the strongest closed models; (3) within a family, confidence quality scales monotonically with capability (Claude: Opus $>$ Sonnet $>$ Haiku), but across families parameter count is a poor predictor; and (4) calibration quality spans a wide range, from near-perfect (Claude Opus) to severely overconfident (Gemma~3-12B), a spread that a per-model post-hoc correction can absorb where absolute values feed a fixed routing threshold, rescaling the scores without changing how they rank extractions. Among logprob aggregations on open-weight models, first-token consistently dominates margin and mean.}

\section{Related Work}
\label{sec:related}

\textbf{Confidence Calibration.} The alignment between a model's predicted confidence and its true probability of correctness is a critical requirement for accuracy-sensitive applications, where confidence scores gate routing decisions between automation and human review.
~\citet{guo2017calibration} demonstrated that modern neural networks are poorly calibrated and tend toward overconfidence, a miscalibration rooted in optimization objectives that prioritize
discriminative performance over probabilistic accuracy. Post-hoc methods such as Temperature Scaling~\cite{guo2017calibration} and adaptive variants~\cite{joy2023sample,balanya2024adaptive} address this for
classification tasks, but do not transfer directly to IDP. For black-box LLMs, ~\citet{tian2023just} showed that verbalized confidence with multiple answer candidates improves calibration over
conditional probabilities. ~\citet{shrivastava2024llamas} introduced surrogate confidence modeling
for black-box predictions, and ~\citet{li2024think} showed that multi-answer reflection reduces overconfidence. Recent work comparing verbalization and token-level log-probability confidence
methods~\cite{xiong2024can,ni2024are,ma2025estimating} finds that neither approach consistently dominates across tasks~\cite{ni2024are}, though token probabilities can be more robust to sampling
temperature~\cite{xie2024calibrating}. For VLMs, temperature scaling improves calibration~\cite{tu2024empirical}, and semantic perturbation can address verbalized
miscalibration~\cite{zhao2025object}. ~\citet{xie2024survey} survey
these approaches comprehensively. While promising for general QA, their application to IDP with multimodal inputs, entity-level predictions, and document degradation remains unexplored.

\begin{table*}[!t]
  \small
          \centering
      \scalebox{0.95}{
      \makebox[\textwidth]{
      \centering
      \begin{tabular}{@{}lcccc@{}}
          \toprule
          \textbf{Dataset} &
          \textbf{Purpose} &
          $\mathbf{\#p}$ &
          \textbf{Color depth} & \textbf{Augmentation} \\ \midrule
      NIST~\cite{dimmick1992nist}             & $f_s$                            & 5590                  & Grayscale &
  \xmark\\
      MARG~\cite{long2005image}            & $f_s$                            & 1553                  & RGB &
  \xmark\\
          Tobacco-800~\cite{zhu2007automatic}             & $f_s$                            & 800                  &
  Grayscale & \xmark\\
           TAB~\cite{mungmeeprued2022tab}             & $f_s$                            & 44.8K                  &
  Grayscale & \xmark\\
    Tobacco-3482~\cite{kumar2013unsupervised}             & $f_p$                            & 3482                  &
  Grayscale & \xmark\\
          RVL-CDIP~\cite{harley2015evaluation}                 & $f_p$ & 400K                  &
  Grayscale & \xmark\\
           \ds~\cite{islam2026docsplit} & $f_{dp}$, $f_p$, $f_d$ & ~1.55M & Mixed & \xmark\\
                RealKIE-FCC-Verified   \cite{amazonagi2024realkiefcc}    & $f_{dp}$, $f_p$, $f_d$, $f_k$  & 383 & Mixed
   & \xmark\\
                  \textbf{{\cbench}} & $f_{dp}$, $f_p$, $f_d$, $f_k$, $f_c$  & 6.7K & RGB &\cmark \\
          \bottomrule
      \end{tabular}
  }}
  \caption{Comparative statistics of   document classification datasets. We present dataset characteristics including
  intended purpose ($f_{dp}$: document packet splitting, $f_s$: form segmentation, $f_p$: form processing, $f_d$: form
  detection, $f_k$: key information extraction, $f_c$: confidence estimation), corpus size ($\#p$: pages), and image
  specifications. }
  \label{tab:dc}
\end{table*}

\textbf{Document Intelligence Benchmarks.} Modern benchmarks evaluate optical character recognition (OCR), classification, KIE, visual question answering (VQA), and table extraction.
RealKIE~\cite{townsend2024realkie} provides five enterprise KIE datasets with PDF documents, OCR output, and text span annotations, though annotation quality varies across the five subsets and may require
manual verification for use as ground truth. UniKIE-Bench~\cite{ji2026unikie} extends to constrained and open-category KIE, and OmniDocBench~\cite{ouyang2025omnidocbench} targets diverse PDF parsing. These
benchmarks focus on accuracy metrics over clean, high-quality documents without real-world degradation artifacts, making them unsuitable for confidence calibration evaluation.

\section{Benchmark Design}
\label{sec:benchmark}

We build upon the RealKIE-FCC-Verified dataset~\cite{amazonagi2024realkiefcc} which is a manually corrected and re-annotated subset of the RealKIE benchmark \cite{townsend2024realkie}. 
The original dataset comprises 75 real-world FCC-format invoice documents with manually verified ground truth annotations. Each document contains multiple entity types (invoice number, date, and line items consisting of five sub-fields), yielding a rich set of extraction targets for calibration analysis. The original documents are production-quality scans with minimal degradation, producing predominantly high-confidence, high-accuracy predictions from IDP pipelines. We augment this dataset with controlled synthetic degradation simulating real-world artifacts.

\revision{\textbf{Licensing and Provenance.} The underlying documents are broadcast political-advertising invoices from the public inspection files that U.S. stations file with the Federal Communications Commission (FCC), and RealKIE-FCC-Verified is released under CC-BY-NC 4.0, which permits adapting and redistributing the material for non-commercial use with attribution. We \releaseverb{} {\cbench} under the same license (\S\ref{sec:appendix_dataset_release}).}

\subsection{Artifact Augmentation}
\label{sec:noise_injection}

To populate the full confidence spectrum, we apply controlled synthetic degradation using Augraphy~\cite{groleau2023augraphy}, an open-source document-specific augmentation framework that simulates realistic physical and digital artifacts. We generate 20 augmented variants per original document, yielding 1,500 augmented documents initially. After quality filtration, 1346 documents remain. For sample document visualization by degradation severity, please refer to \textbf{Appendix~ \ref{sec:appendix_augmentation_examples}}.

\textbf{Augmentation Pipeline Design:} We design 20 augmentation pipelines, 8 pre-configured Augraphy archetypes and 12 custom pipelines targeting specific degradation patterns. Each pipeline applies     
  sequential noise operations across three processing phases: (1)~\textbf{Ink-phase}: ink-level degradation (bleed-through, mottling, low ink, dithering), (2)~\textbf{Paper-phase}: substrate degradation (staining, watermarks, colored paper, moiré patterns, bindings), and (3)~\textbf{Post-phase}: capture and digitization artifacts (JPEG compression, shadows, lighting gradients, scanner defects, geometric   
  distortion). Configurations for the 20 augmentation pipelines are detailed in the \textbf{Appendix~\ref{sec:aug_pipeline_details}}.

\revision{\textbf{Design Intent: Spectrum Coverage over Distribution Fidelity.} The suite spans diverse degradation mechanisms so that calibration can be assessed across the full range of document-quality conditions; it is not intended to reproduce the degradation distribution of the FCC source corpus. It therefore mixes common scanning artifacts, such as JPEG compression, skew, uneven lighting, and photocopier noise, with less frequent degradations that the document-image-processing literature uses to stress-test document understanding systems. Some of the latter, in particular bleed-through, letterpress, and fax artifacts, are implausible for modern single-sided FCC invoices, but they induce controlled accuracy degradation in regions of the spectrum that plausible-only artifacts cannot reach. Without them the low-accuracy end stays too sparsely populated to measure calibration reliably, which is the limitation of the undegraded corpus shown in Figure~\ref{fig:accuracy_histogram}.}

 \textbf{OCR-Safe Parameter Selection:} We constrain augmentation parameters to preserve OCR readability while           
  introducing measurable quality degradation. Key constraints include: (1)~rotation limited to $\pm$~5\textdegree\ as AWS       
  Textract degrades beyond $\pm$10\textdegree, (2)~JPEG quality 70--95 because quality $<$~60 causes severe text degradation,      
  (3)~brightness 0.7--1.3 as values outside this range impair character recognition, (4)~blur radius $\leq$200 pixels with $<$~8   
  iterations, and (5)~overlay alpha $<$~0.3 for text-overlapping effects such as shadows and bleed-through.
  We select pipeline types and noise strengths through an iterative quality assurance process.\newline
 \textbf{Annotation Process and Quality Assessment.} We preserve ground truth annotations from the original RealKIE dataset across all augmented variants, enabling direct comparison of extraction accuracy under varying document quality. Quality assessment ensures that varying pipelines and noise strengths produce a well-distributed spectrum of accuracy degradation, and it excludes documents where augmentation renders them unreadable with catastrophic OCR failure.  
Documents where $>80\%$ of entity types yield null extraction outputs despite having non-null expected values are flagged as catastrophic processing failures. This signature is characteristic of extreme OCR degradation, e.g., \texttt{custom22} dithering pipeline, rendering the document illegible for IDP. Including such documents would artificially inflate calibration error without reflecting genuine model uncertainty. We tune noise-levels to keep the unreadable documents to below $<5\%$ of the dataset while maintaining instances of high degradation.\newline
\textbf{Document Degradation Spectrum Validation:} Each augmentation pipeline applies noise of varying type and strength across 
ink, paper, and post-processing phases. \textbf{Appendix~\ref{sec:aug_pipeline_details}} details the effect and per-phase noise characteristics, including noise strength, for all 20 pipelines. Stronger noise strength and more degradation phases should manifest as higher entity-extraction accuracy drop, the observable 
we use to categorize pipelines by severity. To quantify this, we identify the matched subset of documents present in both the original and augmented 
sets across all nine runs, then compute the per-augmentation accuracy drop. \textbf{Appendix~\ref{sec:appendix_accuracy_drop}} presents 
the accuracy impact by augmentation type; Table~\ref{tab:degradation_tiers} summarizes the three resulting tiers with representative examples. 

\subsection{Dataset Statistics}

Table~\ref{tab:dc} summarizes the final benchmark composition after generation and quality filtration, and provides a comparison against common existing benchmarks. Our benchmark, \textbf{{\cbench}}, comprises 1,346 documents. Entity counts range from 70K to 130K depending on the run, as extraction models differ in output efficiency under fixed token budgets and the fraction of null confidence scores varies by estimation method.

\begin{figure}[!h]
  \centering
  \includegraphics[width=\linewidth]{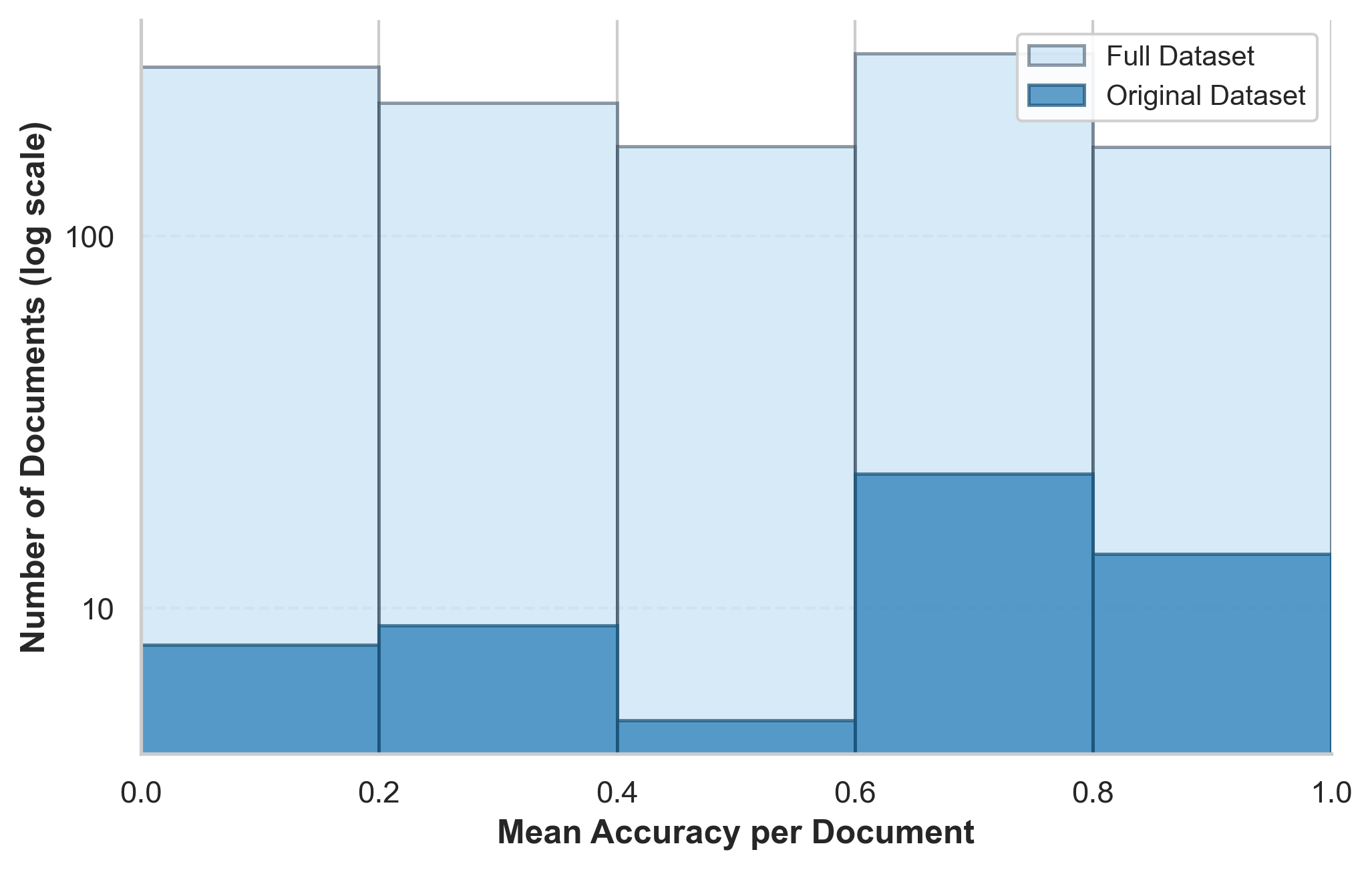}
  \caption{Distribution of per-document mean entity-extraction accuracy (mean over all entity-level binary accuracies per document). (a)~Original dataset with 75 documents concentrate above 0.6, leaving low-accuracy region sparsely populated. (b)~After augmentation, degraded variants populate the 0.0--0.6 range, enabling calibration assessment across the full accuracy range.}
  \label{fig:accuracy_histogram}
\end{figure}

Figure~\ref{fig:accuracy_histogram} confirms the augmentation achieves its intended effect. We compute each document's mean accuracy by averaging its entity-level accuracies, proxying the degree of degradation introduced. The original 75 documents cluster above 0.6, leaving lower accuracy region too sparse for reliable calibration. The augmentation provides statistically sufficient samples across the full spectrum.

\section{Calibration of VLM Models on the Benchmark Dataset}
\label{sec:foundation_models}

We design a systematic evaluation to study how well VLMs produce calibrated confidence measures for KIE. 
Our experimental design spans two axes: (1) two confidence estimation approaches (verbalized and logprob), and (2) three input modality configurations, 
yielding six distinct experimental conditions. We evaluate seven foundation models for the verbalized approach and two open-weight models for the logprob approach, 
enabling controlled comparison of calibration behavior across model families, estimation methods, and input modalities. \textbf{Appendix~\ref{sec:appendix_pipeline}} details the full workflow from optical character recognition (OCR) through classification, information extraction, confidence estimation, evaluation, and calibration assessment.  


\subsection{Experimental Setup}
\label{sec:strategies}
\begin{figure*}[t]
\centering
\includegraphics[width=0.9\textwidth]{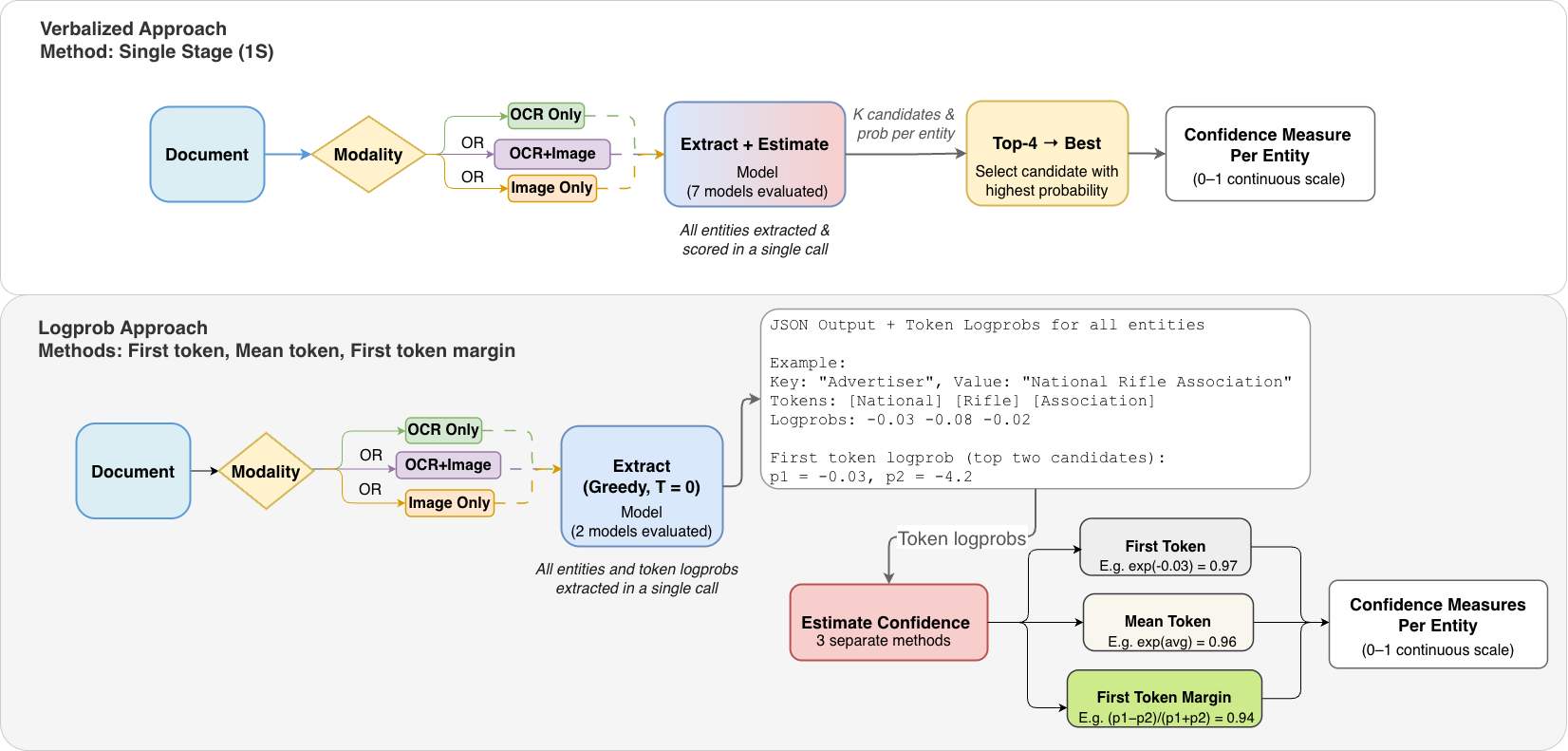}
\caption{Experimental design overview. \textbf{Verbalized:} A single model jointly extracts entities and estimates confidence in one call, producing top candidate guesses with continuous probabilities (0 to 1) per entity; the highest-probability candidate is selected. \textbf{Logprob:} A single-stage extraction pass produces structured output via greedy decoding, and confidence is derived from the token-level log-probabilities of each entity.}
\label{fig:experimental_design}
\end{figure*} 

We study two complementary approaches to confidence estimation for document entity extraction.

\textbf{Verbalized confidence} elicits self-assessed scores directly from the model via prompt engineering. This approach is model-agnostic: it works with any instruction-following VLM and requires no access to internal model states.

\textbf{Logprob confidence} derives scores from token-level log-probabilities produced during greedy decoding. Because it requires access to the raw probability distribution over the vocabulary, this approach is restricted to open-weight models. It requires no prompt engineering and is complementary to verbalized confidence: where verbalized scores depend on the model's ability to introspect, logprob scores capture the model's distributional uncertainty directly.

\textbf{Verbalized method: 1S-TopK.} We employ a single-stage design in which extraction and confidence estimation occur in one LLM call (Figure~\ref{fig:experimental_design}, top). Following the TopK Verbalized approach~\cite{tian2023just}, the model emits its top four candidate guesses with continuous probabilities per entity, and we select the highest-probability candidate. Following~\citet{tian2023just}, we prompt for guesses and probabilities only and omit chain-of-thought reasoning. We evaluate seven VLMs spanning multiple capability tiers and providers: Opus~4.6, Sonnet~4.5, and Haiku~4.5; Kimi~K2.5; Qwen~3~VL-235B and Qwen~3.6-27B; and Gemma~3-12B. 

\textbf{Logprob methods.} Extraction and confidence are obtained from a single greedy decoding pass ($T{=}0$): the model emits the JSON output, and we derive confidence from the logprobs of the tokens covering each entity value (Figure~\ref{fig:experimental_design}, bottom). Let $\{\ell_1,\dots,\ell_K\}$ be the token logprobs spanning a value and $p_1 \ge p_2$ the top-2 probabilities at the first position; we compare three aggregations: \emph{first-token} $\exp(\ell_1)$, \emph{mean token} $\exp(\frac{1}{K}\sum_k \ell_k)$, and \emph{first-token margin} $(p_1{-}p_2)/(p_1{+}p_2)$. Further details are in Appendix~\ref{sec:appendix_logprob_details}. We evaluate two open-weight VLMs for this approach: Qwen~3.6-27B and Gemma~3-12B. \revision{Logprob confidence requires token-level log-probabilities during decoding, which are available only when the model is self-hosted; the parameter count of Qwen~3~VL-235B made self-hosting computationally prohibitive under our infrastructure constraints, so it is evaluated only under the verbalized approach, which is API-compatible.}

\textbf{Input modality configurations:} For each strategy, we evaluate three input configurations: (1)~\textbf{OCR Only}, providing the OCR text extracted by Amazon Textract along with its per-token confidence scores; (2)~\textbf{OCR + Image}, providing both OCR text (with Textract confidence scores) and the document image; and (3)~\textbf{Image Only}, providing only the document image. This yields the factorial design: \{Verbalized, Logprob\} $\times$ \{OCR Only, OCR+Image, Image Only\}.

\revision{\textbf{Scope of comparisons.} Each model's output token budget is held constant across both confidence approaches, so within-model comparisons between strategies are fully controlled. Across models the caps differ by design---40K tokens for Anthropic models, 8K for open-weight models, which degenerate into repetition rather than emit more valid entities when given more room---so cross-model comparisons are only partially controlled, and evaluated entity counts vary with them. All confidence metrics are computed per entity and then aggregated, so the differing set sizes do not bias the scores, but per-model estimates rest on different numbers of entities and we avoid reading small gaps as strict orderings (Appendix~\ref{sec:appendix_logprob_details}).}

\subsection{Evaluation Metrics}
\paragraph{Extraction accuracy.} We report extraction quality via \textbf{Weighted Overall Accuracy (WOA)}, the weighted average of per-field similarity scores across all entity types. For string fields, similarity is the normalized Levenshtein similarity between predicted and ground-truth values; for numeric fields a tolerance-based comparator is used. Each field receives a continuous score between 0 (completely wrong) and 1 (exact match), and WOA is the mean of these scores across all evaluated fields. In this analysis, all entity fields are assigned equal weight. \revision{We additionally report F1, which scores the same predictions on a different scale: WOA awards partial credit through the continuous per-entity similarity score, whereas F1 applies a binary per-field acceptance threshold and counts only exact or near-exact matches. A model can therefore reach a WOA near $0.75$ while its F1 sits near $0.42$; the gap follows from the metric definitions and is not evidence that the task is failing.}

For confidence calibration, we employ four complementary metrics.

\paragraph{AUROC.} AUROC measures discriminative ability: the area under the ROC curve for separating correct from incorrect predictions~\cite{shrivastava2024llamas, xiong2024can}, with random at $0.5$ and perfect at $1.0$. AUROC is scale-invariant.

\paragraph{ECE.} ECE measures calibration error: the weighted absolute gap between predicted confidence and observed accuracy, $\text{ECE}=\sum_{m=1}^{M}\tfrac{|B_m|}{N}\bigl|\overline{\text{acc}}(B_m)-\overline{\text{conf}}(B_m)\bigr|$, computed with adaptive (quantile-based) binning at $M{=}5$ following~\citet{shrivastava2024llamas}; ECE does not capture discriminative ability. \revision{The two are complementary and either can look good while the other looks bad: a model reporting $0.99$ on every correct prediction and $0.90$ on every incorrect one separates the classes perfectly (AUROC $1.0$) while staying badly overconfident on its errors (large ECE), whereas one reporting $0.70$ on every prediction in a set that is $70\%$ correct is calibrated on average (ECE $\approx 0$) yet ranks no better than chance (Appendix~\ref{sec:appendix_ecarb}).}

\paragraph{Brier Score.} The Brier score $\text{BS}=\tfrac{1}{N}\sum_{i=1}^{N}(\text{conf}_i-\text{acc}_i)^2$~\cite{BrierScore_forecastverification} penalizes confident incorrect predictions more heavily than confident correct ones and complements AUROC and ECE without binning artifacts.

\paragraph{Error Capture at Review Budget (ECARB).} A major concern for information extraction deployments is the human effort required to catch errors and raise performance to a target quality threshold. We introduce ECARB to measure the lift of confidence-guided review over random sampling.
Let predictions sorted by ascending confidence, and let $E$ be the total error count. For a review budget
$b \in (0, 1]$, let
$k = \min(n, \max(1, \lfloor bn \rfloor))$ be the number
of predictions reviewed; reviewing $k$ random predictions
catches $(k/n) \cdot E$ errors in expectation:
\begin{equation}
    \text{ECARB}@b \;=\;
\frac{\sum_{i=1}^{k}
\mathbf{1}[\text{acc}_{i} = 0]}{(k/n) \cdot E}
\end{equation}
A value of $1$ indicates no lift over random; higher
$\text{ECARB}@b$ values identify models whose confidence
scores deliver the greatest HITL utility.
\revision{We report $b{=}30\%$ throughout as a representative operating point, not as a recommended review budget; because ECARB depends on the ranking confidence induces rather than on absolute confidence values, model rankings are largely stable across budgets. Appendix~\ref{sec:appendix_ecarb} works the metric through on an example.}


\begin{figure}[!t]
  \centering
  \includegraphics[width=\columnwidth]{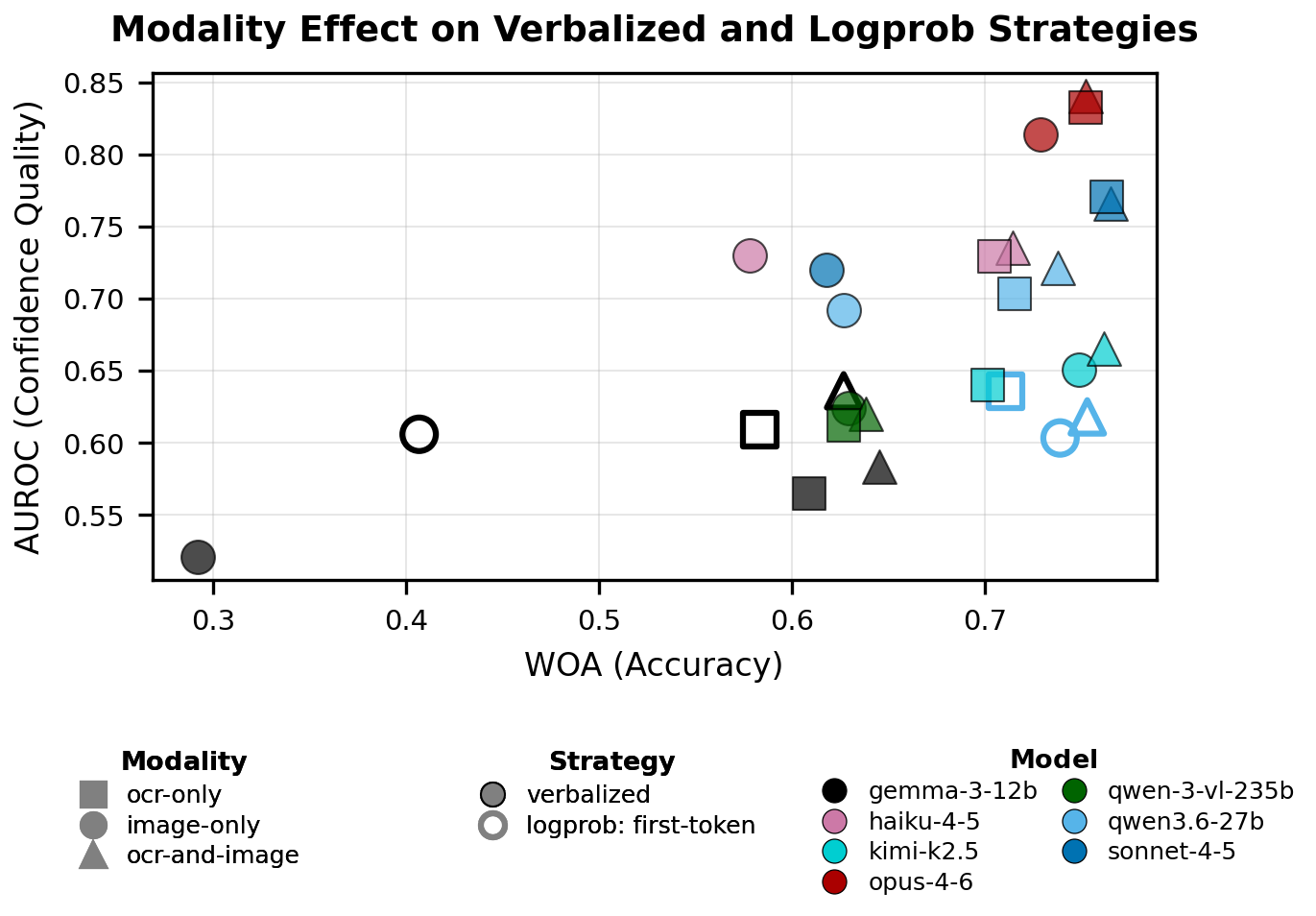}
  \caption{Confidence quality (AUROC) vs.\ extraction accuracy (WOA) across all three input modalities for both verbalized and logprob first-token strategies. OCR+Image consistently occupies the upper-right region, with OCR-only ranking second and image-only third across all models.}
  \label{fig:modality}
\end{figure}

\section{Results and Discussion}\label{sec:results}
The following analysis isolates three factors that influence extraction accuracy and confidence measure: input modality, confidence strategy, and model capability. Across all scatter plots, we use a consistent visual encoding: color denotes model, shape denotes modality (square = OCR, circle = Image, triangle = OCR+Image), and fill style denotes strategy (filled = verbalized, hollow = logprob first-token, $\odot$ = mean-logprob, $\otimes$ = margin).

\subsection{Input Modality}\label{sec:results-modality}

Textual OCR text with OCR confidence scores provides a stronger confidence signal than visual input alone, and combining both modalities is uniformly best (Figure~\ref{fig:modality}). The gap between OCR+Image and image-only ranges from marginal in high-capability models to substantial in weaker ones---Gemma~3-12B drops 6 AUROC points while Opus 4.6 drops only 3. OCR-only occupies a middle position: stronger than image-only but weaker than the combined input. This pattern is independent of confidence strategy, motivating OCR+Image as the standard modality for subsequent analysis. \revision{The practical impact is reflected in ECARB: the strongest OCR+Image configuration (Opus~4.6) yields the highest error-capture gain observed anywhere in the study (Table~\ref{tab:results-ocr-image}).}

\begin{figure}[!t]
  \centering
  \includegraphics[width=\columnwidth]{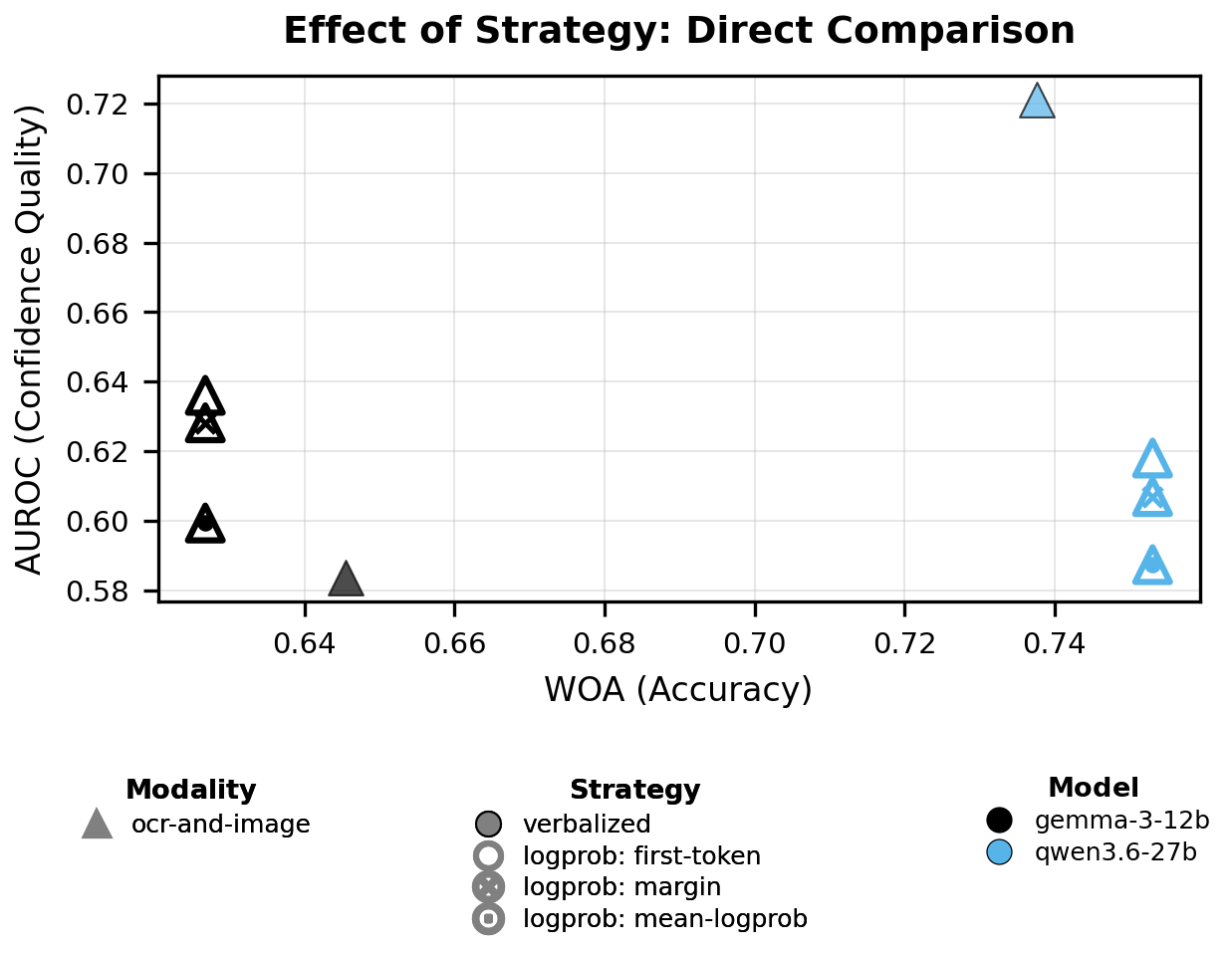}
  \caption{Effect of confidence strategy on AUROC (OCR+Image), restricted to models with all four strategies available. Verbalized outperforms all logprob methods for Qwen~3.6-27B; the pattern reverses for Gemma-3-12B. Among logprob variants, first-token consistently achieves the highest AUROC.}
  \label{fig:strategy}
\end{figure}

\subsection{Confidence Strategy}\label{sec:results-strategy}

Holding modality constant at OCR+Image, confidence strategy has a substantial effect on AUROC, \revision{though the two models that support both families disagree on which strategy wins} (Figure~\ref{fig:strategy}). For Qwen~3.6-27B, verbalized confidence yields 0.72 AUROC compared to 0.62 for logprob first-token, a 10-point advantage. The pattern reverses for Gemma~3-12B, where logprob first-token (0.64) outperforms verbalized (0.58), \revision{consistent with the weaker model being unable to self-assess reliably while still producing informative logit distributions. Extraction accuracy (WOA) differs only marginally and inconsistently between the two strategies---Qwen~3.6-27B moves from 0.74 verbalized to 0.75 logprob while Gemma~3-12B moves from 0.65 to 0.63---so the large AUROC gaps reflect confidence quality rather than task performance.} Among logprob variants, first-token consistently outperforms margin and mean aggregation for both models, making it the preferred logprob method. \revision{Two models cannot establish a general relationship between capability and the preferred confidence family, so we treat this as model-dependent and to be settled per model on held-out data (Appendix~\ref{sec:appendix_analysis_notes}).} In terms of HITL utility (Table~\ref{tab:results-ocr-image}), verbalized Qwen~3.6-27B achieves ECARB=1.93 compared to \revision{1.41} for logprob first-token, confirming that the AUROC advantage translates directly into more errors surfaced per review dollar.

\revision{WOA is not identical across strategies because logprob confidence is read post-hoc off a standard greedy-decoding pass, whereas verbalized 1S-TopK extracts and scores jointly, letting the confidence task influence what gets extracted; Appendix~\ref{sec:appendix_logprob_details} details this coupling and why isolating its effect requires a decoupled two-stage variant.}

\subsection{Model Scale and Family}\label{sec:results-model}

\begin{figure}[!t]
\small
  \centering
  \begin{subfigure}[t]{\columnwidth}
    \centering
    \includegraphics[width=\columnwidth]{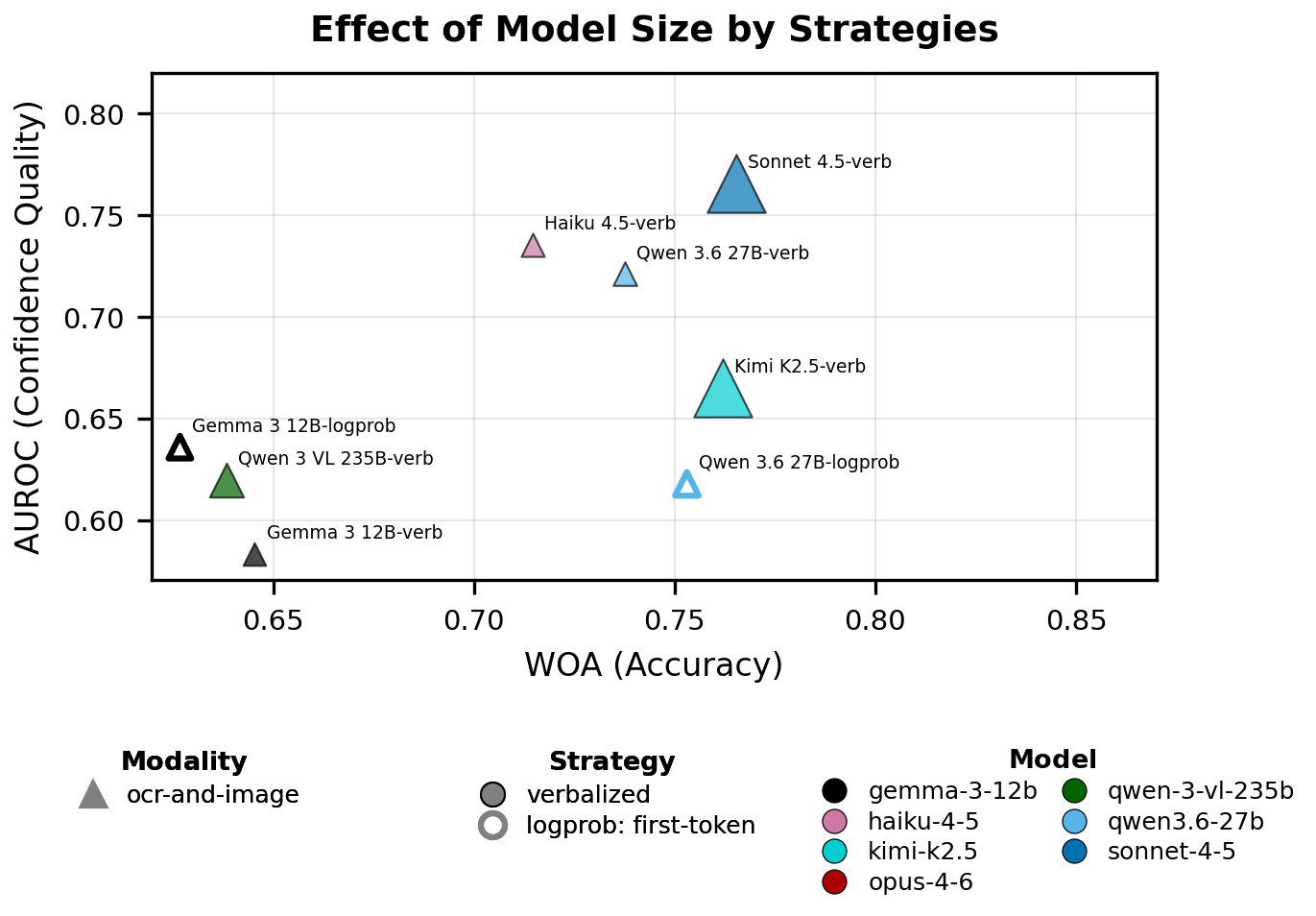}
    \caption{AUROC vs.\ WOA by model. Marker size is proportional to parameter count. Within the Claude family, confidence quality scales monotonically with size.}
    \label{fig:model-size-scatter}
  \end{subfigure}
  \begin{subfigure}[t]{\columnwidth}
    \centering
    \includegraphics[width=\columnwidth]{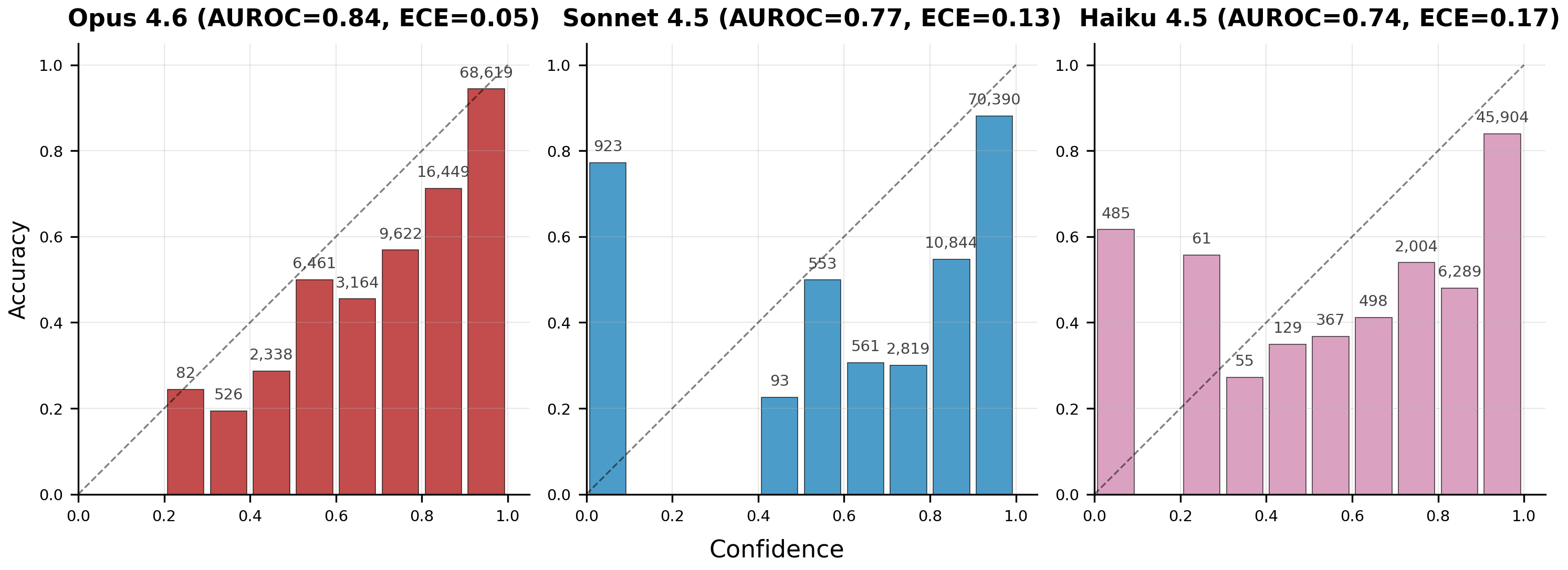}
    \caption{Reliability diagrams for the Claude family. Opus~4.6 (left) closely tracks the y=x diagonal (near-perfect calibration). Sonnet~4.5 (center) and Haiku~4.5 (right) show progressively larger overconfident deviations. Bins with $<$30 samples omitted.}
    \label{fig:model-size-reliability}
  \end{subfigure}
  \caption{Effect of model size and family on confidence quality (OCR+Image). Calibration degrades gracefully across Claude tiers: \revision{Opus~4.6} achieves ECE=0.05 without any post-hoc calibration, while \revision{Haiku~4.5} reaches ECE=0.17.}
  \label{fig:model-size}
\end{figure}

Holding modality to OCR+Image, model capability has the largest effect of the three factors examined (Figure~\ref{fig:model-size}). Within the Claude family, confidence quality scales monotonically with model size: Opus~4.6 achieves the highest AUROC at 0.84, followed by Sonnet~4.5 (0.77) and Haiku~4.5 (0.74). Across families, parameter count is a poor predictor of confidence quality: Qwen~3.6-27B (0.72) surpasses the much larger Qwen~3~VL-235B (0.62) despite having fewer than one-eighth of its parameters, indicating that architecture generation and training methodology are stronger determinants than raw scale. \revision{The two MoE models in our pool (Kimi~K2.5 and Qwen~3~VL-235B) also trail dense models of comparable active parameter count on AUROC, but two models cannot support an architecture-level claim and these models differ from the dense ones in several other respects (Appendix~\ref{sec:appendix_analysis_notes}).} The ECARB metric mirrors this hierarchy: Opus~4.6 achieves 2.43$\times$ gain, Sonnet~4.5 reaches 2.24$\times$, and Haiku~4.5 delivers 1.96$\times$, while the weaker models fall below 1.7$\times$, so model capability directly determines the practical value of confidence-guided review.

The reliability diagrams (Figure~\ref{fig:model-size-reliability}) confirm that the AUROC differences reflect genuine calibration quality\revision{: Opus~4.6 tracks the y=x diagonal across all confidence bins without post-hoc adjustment, while Sonnet~4.5 and then Haiku~4.5 deviate progressively toward overconfidence, suggesting that larger Claude models develop stronger introspective ability during training.} Full reliability diagrams for all models, modalities, and strategies are provided in Appendix~\ref{sec:appendix_reliability}.

\revision{These diagrams also bound what post-hoc calibration can buy: rescaling improves ECE, which matters for absolute-threshold routing, but it is monotone and so leaves AUROC and ECARB essentially unchanged. Calibration is therefore situational rather than a blanket requirement; we do not report calibrated results, and Appendix~\ref{sec:appendix_analysis_notes} explains why.}

\begin{table*}[t]
\centering
\small
\begin{tabular}{llcccccc}
\toprule
Approach & Model & Acc. & F1 & AUROC$\uparrow$ & ECE$\downarrow$ & Brier$\downarrow$ & ECARB$\uparrow$ \\
\midrule
    verbalized & Opus~4.6 & 0.75 & 0.42 & \textbf{0.84} & \textbf{0.05} & \textbf{0.12} & \textbf{2.43} \\
     & Sonnet~4.5 & \textbf{0.77} & 0.51 & 0.77 & 0.13 & 0.16 & 2.24 \\
     & Haiku~4.5 & 0.71 & 0.38 & 0.74 & 0.17 & 0.19 & 1.96 \\
     & Qwen~3.6-27B & 0.74 & 0.41 & 0.72 & 0.22 & 0.23 & 1.93 \\
     & Kimi~K2.5 & 0.76 & 0.46 & 0.67 & 0.20 & 0.22 & 1.68 \\
     & Qwen~3~VL-235B & 0.64 & 0.41 & 0.62 & 0.25 & 0.26 & 1.53 \\
     & Gemma~3-12B & 0.65 & 0.28 & 0.58 & 0.31 & 0.31 & 1.39 \\
\midrule
    logprob (first-token) & Gemma~3-12B & 0.63 & 0.38 & 0.64 & 0.36 & 0.36 & 1.45 \\
     & Qwen~3.6-27B & 0.75 & \textbf{0.70} & 0.62 & 0.21 & 0.21 & 1.41 \\
\midrule
    logprob (margin) & Gemma~3-12B & 0.63 & 0.38 & 0.63 & 0.35 & 0.35 & 1.43 \\
     & Qwen~3.6-27B & 0.75 & \textbf{0.70} & 0.61 & 0.21 & 0.21 & 1.40 \\
\midrule
    logprob (mean-token) & Gemma~3-12B & 0.63 & 0.38 & 0.60 & 0.36 & 0.36 & 1.30 \\
     & Qwen~3.6-27B & 0.75 & \textbf{0.70} & 0.59 & 0.21 & 0.21 & 1.36 \\
\bottomrule
\end{tabular}
\caption{Extraction accuracy and confidence calibration metrics on the \textbf{OCR+Image} modality, selected as it consistently yields the strongest joint performance in both WOA and AUROC across approaches. In this setting, the model receives both the raw document image and the OCR text extracted by Amazon Textract along with its per-token confidence scores. ECARB reports the error-capture gain at 30\% review budget.}
\label{tab:results-ocr-image}
\end{table*}

\subsection{Metrics and Their Implications}\label{sec:results-metrics}

Table~\ref{tab:results-ocr-image} reveals that different metrics capture distinct aspects of confidence quality, and no single model dominates across all dimensions. AUROC and ECE are complementary: Opus~4.6 achieves the best scores on both (0.84 and 0.05, respectively), reflecting both strong discrimination and well-calibrated confidence. In contrast, Sonnet~4.5 ranks second in AUROC (0.77) but has a higher ECE (0.13), indicating that its confidence scores correctly discriminate between correct versus incorrect predictions, but are not well-calibrated in absolute terms. Logprob-based approaches show systematically higher ECE than verbalized methods, despite comparable AUROC in some cases, suggesting that logit distributions are informative for discrimination between correct versus incorrect extractions, \revision{but would need rescaling before being read as absolute confidence values---though not before being used to order a review queue.} Finally, ECARB provides the most practically relevant signal: Opus~4.6 captures 2.43 times as many errors as a random reviewer at a 30\% review budget, while even the weakest logprob approach (mean) yields a 1.30$\times$ gain. \revision{We use 30\% throughout as a representative operating point, not a recommended budget; since ECARB depends only on the ranking confidence induces over predictions, the ordering of configurations is largely stable as the budget varies, and it should be read at whatever budget a deployment can staff.}

\revision{The F1 column sits well below WOA because the two score the same predictions on different scales, not because the task is failing (\S\ref{sec:foundation_models}), and a moderate error rate is in any case the regime in which confidence routing is most useful: were extraction near-perfect, a reviewer would have little left to find. The six columns are also not meant to be weighed equally. ECARB speaks most directly to the deployment question, WOA and AUROC form the diagnostic decomposition behind it, ECE matters only for absolute-threshold routing, and Brier is a proper-scoring-rule sanity check; Appendix~\ref{sec:appendix_ecarb} gives a reading guide and works ECARB through on an example.}

\paragraph{\revision{Practitioner Guidelines.}} \revision{For HITL workflows that review predictions in ascending confidence order, extraction accuracy and confidence quality must be weighed jointly: WOA fixes how many errors a batch contains, while AUROC and ECARB fix what fraction of them review surfaces per unit of effort. The two can diverge sharply---Sonnet~4.5 attains the highest extraction accuracy (0.77 WOA) yet Opus~4.6 leads on every confidence metric, and Kimi~K2.5 reaches competitive accuracy (0.76 WOA) while lagging in AUROC (0.67). We therefore suggest shortlisting on WOA and AUROC together, adding post-hoc calibration only where threshold-based routing is required, and reading ECARB at the deployment's own review budget. Appendix~\ref{sec:appendix_analysis_notes} develops these tradeoffs.}

\section{Conclusion}
\label{sec:conclusion}

This work presents ConfBench and a systematic evaluation of VLM confidence calibration for document KIE. Our results demonstrate that confidence quality is primarily driven by model capability and input modality\revision{; which of the two confidence families scores better is model-dependent, an observation we scope to the two open-weight models that expose token-level probabilities (\S\ref{sec:results-strategy})}. The proposed ECARB metric bridges the gap between statistical calibration measures and deployment economics: at a 30\% review budget, the best configuration surfaces 2.4$\times$ more errors than random sampling, while even the weakest yields 1.3$\times$, confirming that confidence-guided review delivers tangible operational value across all evaluated settings.

\revision{The scope constraints of \cbench, our roadmap for extending it to further domains and source corpora, and the deployment risks of confidence-guided routing are discussed in the Limitations section.}

\section*{Limitations}
\label{sec:limitations}

\paragraph{Scope of the Source Corpus.}
\revision{{\cbench} is built on 75 manually verified FCC-format invoice documents~\cite{amazonagi2024realkiefcc}, drawn from one document domain, one broad layout family, and one language. Our findings therefore describe how the evaluated VLMs behave on degraded variants of business invoices, and we scope them accordingly rather than presenting them as claims about document understanding in general. Confidence quality is a joint property of the model, the entity schema, and the visual structure of the input, so whether our conclusions about modality, confidence strategy, and relative model ranking carry over to other document types, to multi-column or handwritten layouts, to non-English documents, or to degradation arising naturally in a production scanning pipeline is an open question that the extensions below are designed to answer.}

\paragraph{Corpus Size as a Deliberate Tradeoff.}
\revision{The size of the source set reflects a tradeoff we made deliberately in favor of annotation quality. Calibration evaluation is far more sensitive to annotation error than accuracy evaluation is: a mislabeled entity does not simply lower measured accuracy, it moves a sample to the wrong side of the correct/incorrect boundary and thereby corrupts every reliability bin it falls into, biasing ECE, AUROC, and ECARB at once. Public KIE corpora cover many more document types, but annotation quality varies across subsets~\cite{townsend2024realkie}, which makes them unsuitable as calibration ground truth without further curation. We selected the FCC subset because it has been manually corrected and re-annotated, accepting a smaller document count in exchange for entity-level ground truth we can trust. Verified annotation of this kind is labor- and resource-intensive, and it, not the augmentation pipeline, is what binds corpus size.}

\paragraph{Planned Expansion of {\cbench}.}
\revision{We treat {\cbench} as a seed benchmark rather than a finished one, and we intend to maintain and extend it as new models and confidence estimation methods appear (\S\ref{sec:appendix_dataset_release}). Our roadmap has four parts. First, additional document domains and types beyond invoices, covering different entity schemas and layout conventions. Second, additional source corpora: we will investigate DocILE~\cite{simsa2023docile}, which contains roughly 6,500 annotated documents, as a candidate expansion set, contingent on its annotations meeting the reliability bar that calibration evaluation requires; where they do not, targeted re-annotation of a subset is the fallback. Third, broader language coverage, separating confidence behavior from English-specific tokenization and OCR quality. Fourth, naturally occurring degradation collected from real capture and scanning workflows, to complement the controlled synthetic degradation used here rather than replace it. Every extension must preserve the design requirement that motivates the benchmark: samples have to span the full accuracy spectrum, since a corpus concentrated at high accuracy cannot separate a well-calibrated estimator from an overconfident one.}

\paragraph{Augmentation Targets Mechanism Coverage.}
\revision{Our augmentation suite is designed to span a broad range of degradation mechanisms and quality levels so that calibration can be assessed across the accuracy spectrum, rather than to reproduce the degradation distribution of FCC invoices. A few included artifacts, such as bleed-through and fax transmission noise, are uncommon for modern single-sided invoices, yet they remain functionally necessary to populate the low-accuracy region; \S\ref{sec:noise_injection} states the full argument and the readability constraints that bound it. Collecting naturally degraded scans to sit alongside the synthetic variants is part of the planned expansion above.}

\paragraph{Model Coverage Behind Two Observations.}
\revision{Two of our observations rest on small model samples, and we present them as scoped observations rather than general claims. The comparison between verbalized and log-probability confidence is available only for the two open-weight models that expose token-level probabilities (\S\ref{sec:results-strategy}), so the interaction we observe between model capability and the better-performing confidence family is a two-point pattern rather than an established trend. The MoE observation likewise rests on the two MoE models in our evaluation (\S\ref{sec:results-model}); with $n=2$ per architecture class and no control over pretraining data or post-training recipe, we deliberately stop short of attributing the gap to sparse activation itself. Both are hypotheses that a larger model panel can settle, and extending the panel as new open-weight models expose token-level probabilities is part of our maintenance plan for the benchmark (\S\ref{sec:appendix_dataset_release}).}

\paragraph{Degree of Control in Cross-Model Comparisons.}
\revision{Comparisons within a single model across confidence strategies and modalities are fully controlled: the same documents, prompts, and entity schema are used throughout, so differences are attributable to the manipulated factor. Cross-model comparisons are controlled to a lesser degree. Output token budgets differ by provider by design (40K tokens for Anthropic models, 8K for open-weight models), following each provider's serving limits, and the number of evaluated entities varies across models because \texttt{LineItems} is a nested variable-length list whose length depends on how much of the document a model transcribes within its budget (\S\ref{sec:appendix_logprob_details}). We therefore read absolute metric values across models as indicative of relative standing rather than as the output of a strictly matched comparison, and we avoid treating small cross-model gaps as strict orderings.}

\paragraph{Calibration Methods as the Next Use of {\cbench}.}
\revision{The scope of this work is to characterize calibration quality across models, modalities, and confidence families, and to supply the evaluation substrate that comparing calibration methods requires. Benchmarking specific calibrators is the natural next use of that substrate rather than part of the present study, so we report no post-hoc calibration results and make no claim about how much any calibrator would help. Two considerations shape how we plan to approach it. First, post-hoc calibration is situational rather than universally required: it rescales absolute confidence values and matters when routing depends on an absolute threshold, but monotone rescaling leaves the confidence ranking untouched, and with it AUROC and ECARB. Second, where calibrated absolute values are needed, the reliability diagrams in Appendix~\ref{sec:appendix_reliability} already expose a per-bin correction directly. Building on this, we intend to evaluate temperature scaling and comparable calibrators on {\cbench}, including whether a calibrator fitted on one degradation tier transfers to another, a question the benchmark's tiered structure is well suited to answer.}

\paragraph{Additional Future Directions.}
\revision{Two extensions to the confidence estimators themselves follow from our results. A fully decoupled two-stage verbalized approach, in which extraction is completed first and confidence is scored in a separate call, would hold extraction output fixed and isolate confidence quality in the way the log-probability setup already does. Combining verbalized and log-probability signals into an ensemble score is a second direction, since the two families draw on different information and the better of the two varies by model in our results.}

\paragraph{Potential Risks.}
\revision{Confidence-guided routing complements rather than replaces human oversight. Calibration established on one distribution may shift under deployment conditions, and a system that treats high confidence as a substitute for review will silently pass through the errors the model is most certain about. Because {\cbench} measures calibration on synthetically degraded invoices, operators should not carry our thresholds or ECARB figures over to a different document population without re-measuring on their own data. Sustaining trust as {\cbench}-informed systems reach new domains requires ongoing distribution monitoring, periodic recalibration, and a residual audit of automatically accepted extractions so that confident errors stay observable.}


\bibliography{ref}

\clearpage
\newpage
\appendix
\section{Appendix}
\subsection{Dataset Release, Licensing, and Maintenance}
\label{sec:appendix_dataset_release}

\revision{
\paragraph{Release and contents.} We \releaseverb{} {\cbench} publicly as a Hugging Face dataset.\dataseturlinline{} The release \releasecontains{} the 75 original RealKIE-FCC-Verified invoice images together with the 1{,}346 augmented documents that constitute {\cbench} after quality filtration (\S\ref{sec:benchmark}); the ground-truth entity annotations, which are preserved unchanged from the source dataset across every augmented variant, so that each variant is directly comparable to its original; the augmentation pipeline configurations for the 20 pipelines characterized in Table~\ref{tab:augmentation_characteristics} (\S\ref{sec:aug_pipeline_details}), including the ordered per-phase operations and the parameter settings under which they were applied; and per-document metadata recording the source document, the pipeline that produced each variant, and its degradation tier as defined in Table~\ref{tab:degradation_tiers}. These artifacts let others regenerate the augmented corpus from the originals, reproduce our evaluation on the exact images we used, and stratify calibration results by degradation severity.
}

\revision{
\paragraph{License and attribution.} {\cbench} is \releasedstate{} under the Creative Commons Attribution-NonCommercial 4.0 International license (CC-BY-NC-4.0). This choice inherits from the source corpus: RealKIE-FCC-Verified~\cite{amazonagi2024realkiefcc} is itself released under CC-BY-NC 4.0, which permits adaptation and redistribution of adaptations provided that attribution is given and use remains non-commercial. Our augmented variants are adaptations in exactly this sense, so releasing them under the same terms keeps the derived benchmark license-compatible with its source. The dataset card \releaseattributes{} both RealKIE-FCC-Verified and the original RealKIE benchmark~\cite{townsend2024realkie}, and \releaserecords{} Augraphy~\cite{groleau2023augraphy} as the augmentation framework.
}

\revision{
\paragraph{Provenance.} The underlying documents are invoices drawn from the public inspection files that broadcast stations are required to file with the U.S.\ Federal Communications Commission (FCC). They are therefore public records rather than private enterprise documents, which is what makes redistribution of the images possible at all; it is also why we build on this corpus instead of a proprietary invoice collection, despite the originals being production-quality scans that required augmentation to populate the low-accuracy regime (\S\ref{sec:noise_injection}).
}

\ifanonbuild
\revision{
\paragraph{Anonymity constraint on the URL.} We do not include the dataset URL in this submission. A public dataset listing carries owner and organization metadata, so linking to it would create an identifying, de-anonymizing page and violate double-blind review. The URL will be added to the camera-ready version once the review period is over; the release itself is not contingent on the review outcome.
}
\fi

\revision{
\paragraph{Maintenance.} We intend to keep {\cbench} in active maintenance rather than releasing it once and leaving it fixed. We \releasepublish{} versioned releases so that results remain reproducible against a fixed snapshot while the benchmark grows, with each version documenting the models, confidence estimation methods, and documents it covers. We plan to broaden the evaluation over time by adding new VLMs as they are released and additional confidence estimation methods beyond the verbalized and logprob families studied here, and to extend coverage to further document domains beyond invoices, where the entity schema and the dominant failure modes differ. Corrections to annotations reported by users will be folded into subsequent versions with a changelog, so that any change in reported numbers is traceable to a change in the data.
}

\subsection{Augmentation Quality Examples}
\label{sec:appendix_augmentation_examples}

\begin{figure*}[!ht]
\centering
\begin{subfigure}[t]{0.48\textwidth}
    \centering
    \includegraphics[width=\textwidth,height=0.35\textheight,keepaspectratio]{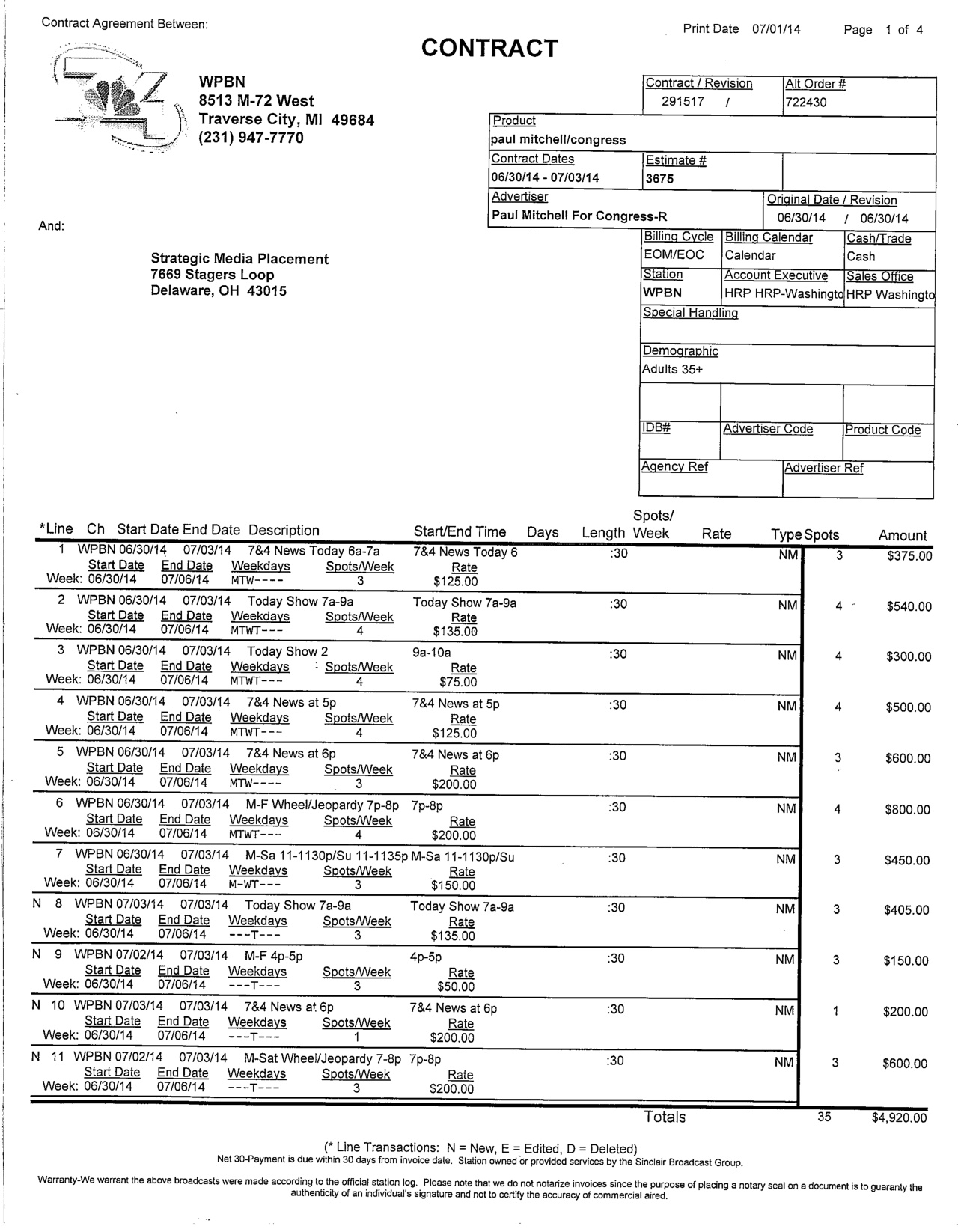}
    \caption{Original}
\end{subfigure}
\hfill
\begin{subfigure}[t]{0.48\textwidth}
    \centering
    \includegraphics[width=\textwidth,height=0.35\textheight,keepaspectratio]
    {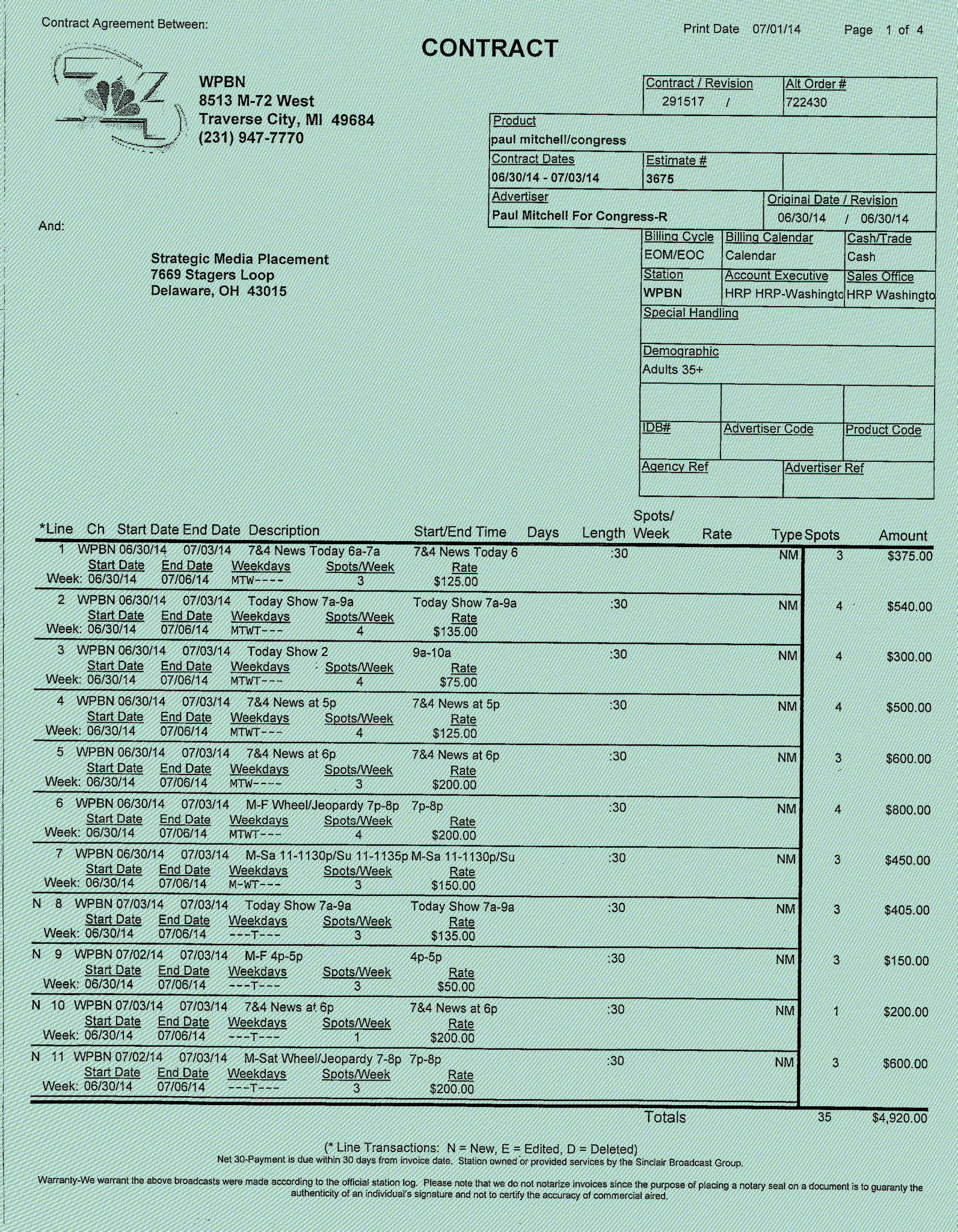}
    \caption{\texttt{Custom15} — mild degradation}
\end{subfigure}

\begin{subfigure}[t]{0.48\textwidth}
    \centering
    \includegraphics[width=\textwidth,height=0.35\textheight,keepaspectratio]{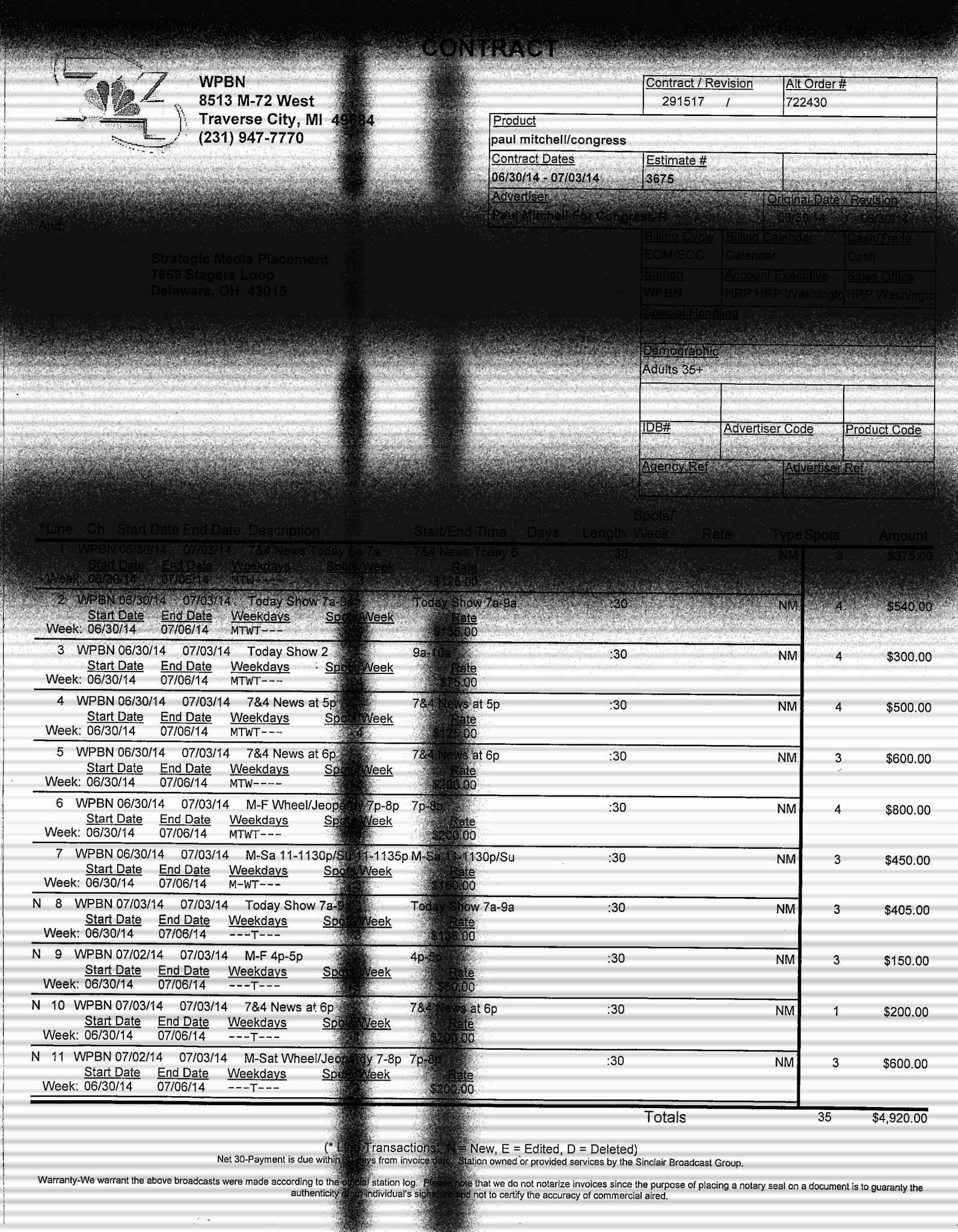}
    \caption{\texttt{Custom12} — moderate degradation}
\end{subfigure}
\hfill
\begin{subfigure}[t]{0.48\textwidth}
    \centering
    \includegraphics[width=\textwidth,height=0.35\textheight,keepaspectratio]{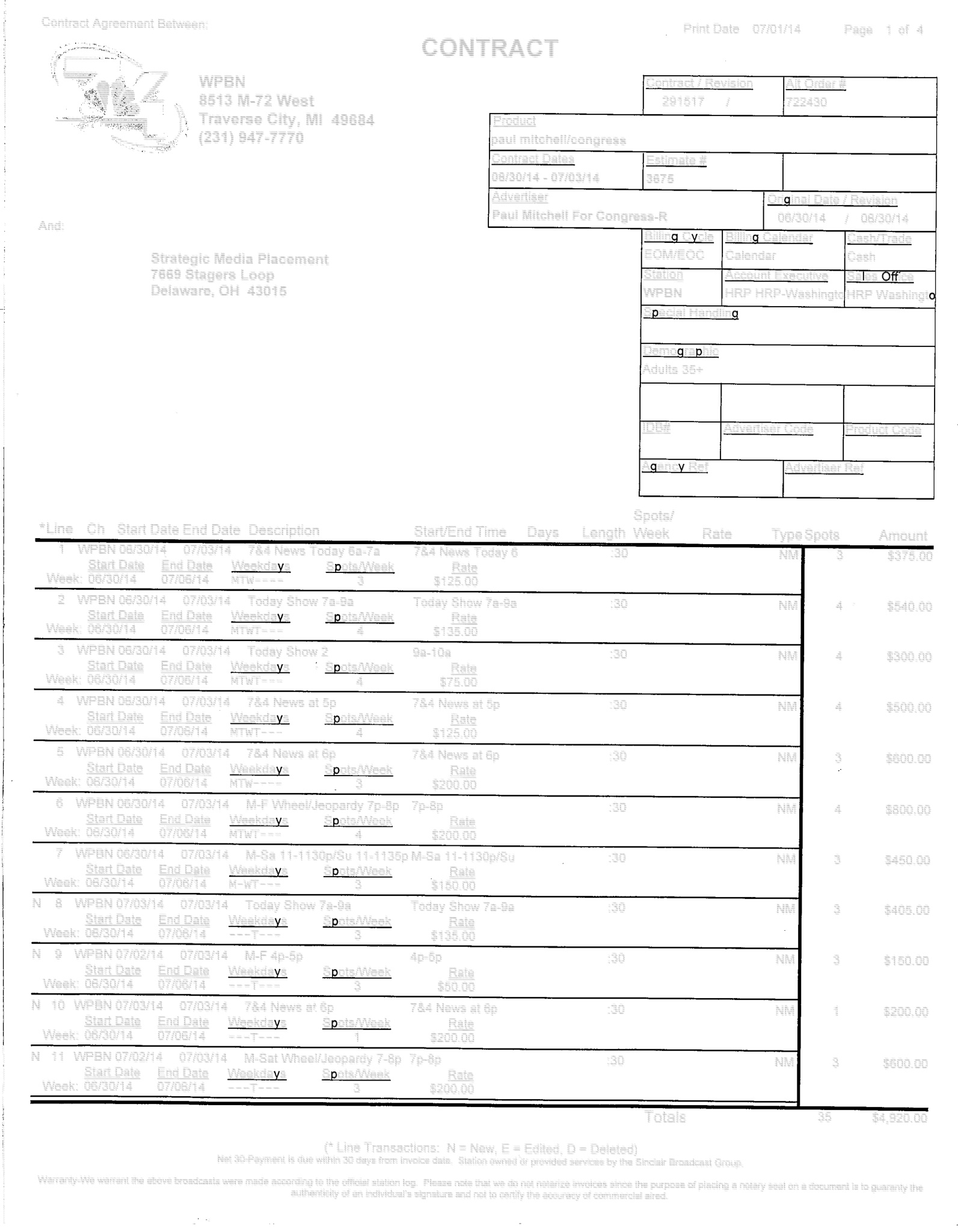}
    \caption{\texttt{Custom22} — high degradation}
\end{subfigure}
\caption{Document \texttt{1220d048d2a75d910c52e7f290e32ca0} from our {\cbench} dataset under three degradation levels. (a)~Original clean scan. (b)~\texttt{Custom15}: mild degradation ($<$5\% accuracy drop) simulates Moir\'{e} pattern on colored paper. (c)~\texttt{Custom12}: moderate degradation (6--10\% accuracy drop) simulates dirty printer drum and rollers. (d)~\texttt{Custom22}: high degradation (21\% accuracy drop) represents dithered print with dot-matrix simulation.}
\label{fig:augmentation_examples}
\end{figure*}

Figure~\ref{fig:augmentation_examples} illustrates a single source document rendered under three representative pipelines, one from each severity tier defined in Table~\ref{tab:degradation_tiers}. The same document is used across all four panels so that visual differences are attributable to the augmentation pipeline rather than to underlying document content. Ground truth annotations are preserved across variants (\S\ref{sec:noise_injection}), making each panel directly comparable in terms of extraction difficulty.

\textbf{Mild tier ($<$5\% accuracy drop).} Panel~(b) shows \texttt{Custom15}, which applies a moir\'{e} pattern over colored paper in the paper phase only (Table~\ref{tab:augraphy_pipelines}). The text remains crisp and character strokes are preserved; the dominant artifact is a low-frequency tonal modulation of the background. OCR confidence on body text drops only marginally and extraction accuracy degrades by less than 5\%, but the colored cast introduces a measurable shift in document-image appearance that is visible to image-only confidence estimators.

\textbf{Moderate tier (5 to 10\% accuracy drop).} Panel~(c) shows \texttt{Custom12}, a post-phase pipeline combining dirty-drum streaks with roller marks. Vertical and horizontal banding overlay the text region, locally occluding glyphs without rendering them globally illegible. Extraction errors concentrate on entities whose bounding boxes intersect heavy banding (e.g., line-item digits, date separators). This is the largest tier by design: it covers the range of degradation most commonly observed in production scans, where individual artifacts are visible but the document remains machine-readable.

\textbf{High tier ($>$10\% accuracy drop).} Panel~(d) shows \texttt{Custom22}, which combines ink-phase dithering with a post-phase dot-matrix simulation. Character interiors are broken into dot patterns and inter-character spacing becomes irregular, pushing OCR toward the boundary of the readability envelope defined by our parameter constraints (\S\ref{sec:noise_injection}): rotation $\leq \pm 5^{\circ}$, JPEG quality $\geq 70$, blur radius $\leq 200$\,px, and overlay alpha $<0.3$. Documents whose augmentation pushes more than 80\% of entity types to null extraction under these constraints are removed by the unreadability filter (\S\ref{sec:noise_injection}), so the surviving high-tier documents represent severe but not catastrophic degradation. This regime is where calibration matters most: extraction accuracy is no longer near-perfect, and confidence scores must reliably separate the remaining correct extractions from the now-frequent incorrect ones.

\textbf{Quality control.} Across all three tiers, two properties are preserved by construction. First, ground truth values remain unchanged, so accuracy degradation reflects model robustness rather than annotation drift. Second, the OCR-safe parameter envelope (\S\ref{sec:noise_injection}) keeps the share of documents removed by the unreadability filter below 5\% of the dataset, so the visible degradation in panels~(b)--(d) is representative of what enters {\cbench} rather than an extreme tail. Per-pipeline configurations underlying these examples are listed in Table~\ref{tab:augraphy_pipelines}.

\subsection{Augmentation Pipeline Characteristics}
\label{sec:aug_pipeline_details}

We employ the Augraphy library to generate realistic document degradation patterns across 20 distinct noise pipelines. Each pipeline is structured as a three-phase transformation process that mimics the physical lifecycle of document degradation: \textit{Ink Phase}, \textit{Paper Phase} and \textit{Post Phase}. We utilize eight pre-defined archetype pipelines from the Augraphy library alongside twelve custom-designed pipelines to achieve diverse noise characteristics. 
Each pipeline is assigned a noise intensity level (High, Medium, or Low) per phase based on the severity and number of augmentations applied, enabling systematic evaluation across varying degradation conditions. This multi-phase approach ensures that our synthetic noisy documents closely replicate real-world    scenarios where documents undergo multiple forms of degradation throughout their lifecycle. Table~\ref{tab:augmentation_characteristics} shows which real-world effect does each augmentation pipeline simulate, and the details of the Ink, Paper and Post phases.

   \begin{table*}[!ht]
   \centering
   \small
   \setlength{\tabcolsep}{3pt}
   \begin{tabular}{@{}lp{3.6cm}p{2.6cm}p{2.6cm}p{2.6cm}@{}}
   \toprule
   \textbf{Pipeline} & \textbf{Real-World Effect} & \textbf{Ink Phase} & \textbf{Paper Phase} & \textbf{Post Phase} \\
   \midrule
   Default & General multi-artifact degradation & \cellcolor{yellow!20}\textbf{M:} InkColorSwap, Dithering, InkBleed, Letterpress & \cellcolor{yellow!20}\textbf{M:} PaperFactory, ColorPaper, WaterMark & \cellcolor{yellow!20}\textbf{M:} DirtyDrum, Jpeg, BadPhotoCopy, Geometric, Faxify \\
   \addlinespace
   Archetype2 & Marked document with punch holes, photocopier noise and reflected glare & \cellcolor{gray!20}\textbf{L:} Markup, InkBleed & \cellcolor{yellow!20}\textbf{M:} Bindings, PageBorder & \cellcolor{red!20}\textbf{H:} BadPhotoCopy, Geometric, Scribbles, ReflectedLight \\
   \addlinespace
   Archetype3 & Heavy ink bleed with lighting gradients & \cellcolor{yellow!20}\textbf{M:} InkBleed & \cellcolor{gray!20}\textbf{L:} BadPhotoCopy & \cellcolor{yellow!20}\textbf{M:} LightingGradient, BleedThrough \\
   \addlinespace
   Archetype4 & Cropped fax with punch holes and monochrome conversion & \cellcolor{red!20}\textbf{H:} Geometric, Bindings, InkBleed & \cellcolor{yellow!20}\textbf{M:} Geometric & \cellcolor{gray!20}\textbf{L:} Faxify, BadPhotoCopy \\
   \addlinespace
   Archetype7 & Minimal fax with moderate ink bleed and background noise & \cellcolor{gray!20}\textbf{L:} Geometric & \cellcolor{red!20}\textbf{H:} BadPhotoCopy & \cellcolor{yellow!20}\textbf{M:} Faxify, InkBleed, BadPhotoCopy \\
   \addlinespace
   Archetype9 & Letterpress with fax processing and noisy lines & \cellcolor{yellow!20}\textbf{M:} Letterpress, Faxify, Dithering & \cellcolor{yellow!20}\textbf{M:} NoisyLines, BadPhotoCopy & \cellcolor{red!20}\textbf{H:} Geometric \\
   \addlinespace
   Archetype10 & Faded letterpress with minimal photocopier noise & \cellcolor{red!20}\textbf{H:} BadPhotoCopy, Letterpress & --- & \cellcolor{yellow!20}\textbf{M:} Faxify \\
   \addlinespace
   Archetype11 & Aged paper with color shift, shadows, and bleed-through & \cellcolor{gray!20}\textbf{L:} LinesDegradation, InkBleed, LowInkLines & \cellcolor{yellow!20}\textbf{M:} PatternGenerator, ColorPaper & \cellcolor{red!20}\textbf{H:} ColorShift, Jpeg, ShadowCast, BleedThrough \\
   \midrule
   Custom12 & Dirty printer drum and rollers & --- & --- & \cellcolor{red!20}\textbf{H:} DirtyDrum, DirtyRollers \\
   \addlinespace
   Custom13 & Coffee-stained and folded document & --- & \cellcolor{yellow!20}\textbf{M:} Stains & \cellcolor{yellow!20}\textbf{M:} Folding \\
   \addlinespace
   Custom14 & Heavy ink bleed-through and mottling & \cellcolor{red!20}\textbf{H:} BleedThrough, InkMottling & --- & --- \\
   \addlinespace
   Custom15 & Moir\'{e} pattern on colored paper & --- & \cellcolor{yellow!20}\textbf{M:} Moire, ColorPaper & --- \\
   \addlinespace
   Custom16 & Shadow and uneven lighting & --- & --- & \cellcolor{yellow!20}\textbf{M:} ShadowCast, LightingGradient \\
   \addlinespace
   Custom17 & Heavy JPEG compression with noise & --- & --- & \cellcolor{red!20}\textbf{H:} Jpeg, SubtleNoise \\
   \addlinespace
   Custom18 & Rotated scan & \cellcolor{yellow!20}\textbf{M:} Geometric & --- & \cellcolor{gray!20}\textbf{L:} PageBorder \\
   \addlinespace
   Custom19 & Binding marks with letterpress imprint & --- & \cellcolor{yellow!20}\textbf{M:} Bindings & \cellcolor{red!20}\textbf{H:} Letterpress \\
   \addlinespace
   Custom20 & Extreme brightness with bad photocopy & \cellcolor{red!20}\textbf{H:} Brightness & \cellcolor{red!20}\textbf{H:} BadPhotoCopy & --- \\
   \addlinespace
   Custom21 & Watermarked document with noisy lines & --- & \cellcolor{yellow!20}\textbf{M:} WaterMark & \cellcolor{yellow!20}\textbf{M:} NoisyLines \\
   \addlinespace
   Custom22 & Dithered print with dot-matrix simulation & \cellcolor{yellow!20}\textbf{M:} Dithering & --- & \cellcolor{yellow!20}\textbf{M:} DotMatrix \\
   \addlinespace
   Custom23 & Aged paper stock & --- & \cellcolor{yellow!20}\textbf{M:} PaperFactory & \cellcolor{red!20}\textbf{H:} Faxify \\
   \bottomrule
   \end{tabular}
   \caption{Augmentation pipeline configurations. The \textit{Real-World Effect} column describes the physical scenario each pipeline simulates. Phase columns list noise components with cell color and label indicating severity: \colorbox{red!20}{\textbf{H}igh}, \colorbox{yellow!20}{\textbf{M}edium}, \colorbox{gray!20}{\textbf{L}ow}.}
   \label{tab:augraphy_pipelines}
   \label{tab:augmentation_characteristics}
   \end{table*}

\subsection{Accuracy Drop by Augmentation Type}
\label{sec:appendix_accuracy_drop}

\begin{figure}[h]
\centering
\includegraphics[width=\columnwidth]{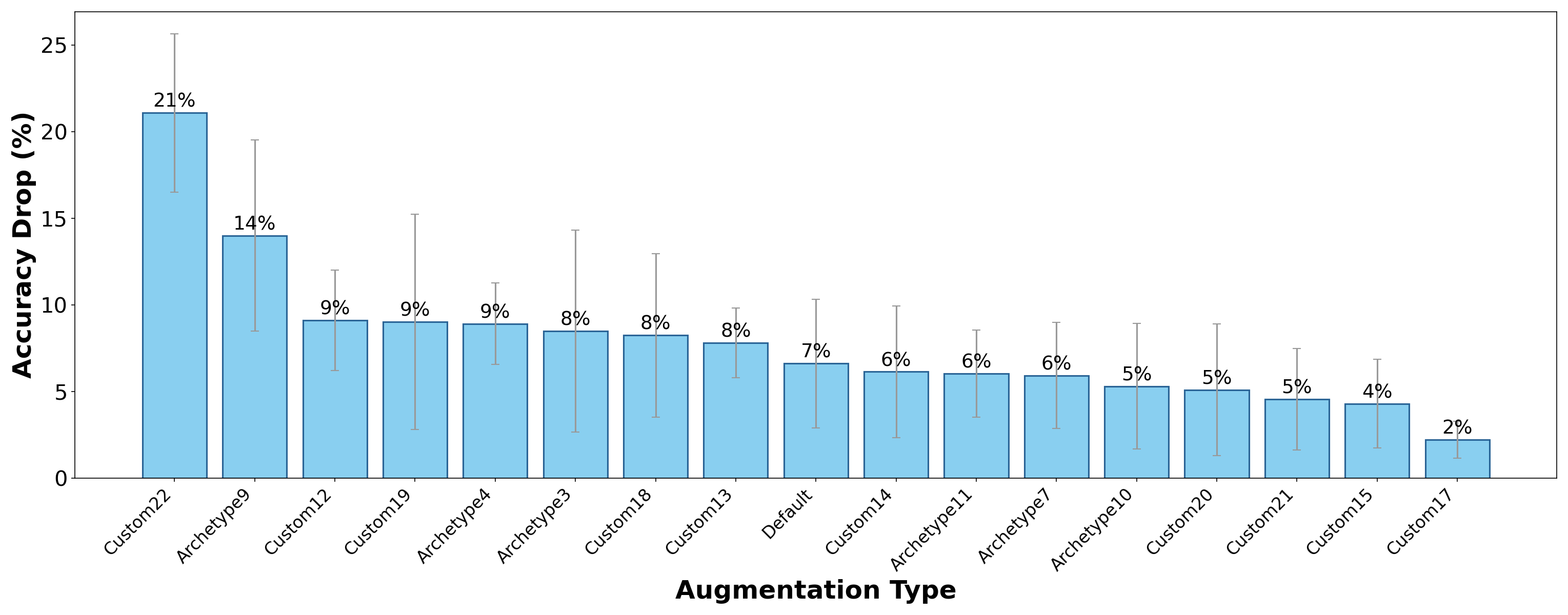}
\caption{Average accuracy drop in information extraction per augmentation type, averaged across nine runs. All runs use Sonnet~4 
as the entity extractor. Error bars show standard deviation across runs. 17 of 20 augmentation types have $N > 1{,}000$ entities per run and mean accuracy drop 
exceeding the cross-run standard deviation. We exclude the 3 augmentation types where run-to-run variability obscures a consistent degradation signal. Types are sorted by decreasing severity.}
\label{fig:accuracy-drop}
\end{figure}

For each augmentation type, we compute the accuracy drop as the difference in mean entity-level extraction accuracy between the original documents and their augmented counterparts, 
using only the matched subset of documents present in both sets. A higher accuracy drop indicates that the augmentation more severely degrades extraction accuracy. The full per-pipeline accuracy-drop distribution is shown in Figure~\ref{fig:accuracy-drop}.

\subsection{Document Processing Pipeline}                            \label{sec:appendix_pipeline} 


\textbf{Optical Character Recognition (OCR).} We use Amazon Textract with the \textsc{Layout} feature to extract text and structural elements from each document image. Textract returns a word-level output: each token carries a detected string, an axis-aligned bounding box, a reading-order index, and a per-token OCR confidence score in $[0, 100]$. When the OCR modality is active, we serialize this output as a structured text block of \texttt{(word, confidence)} pairs, preserving reading order, and pass it in the user message. For the \emph{OCR-Only} modality the document image is withheld entirely; for \emph{OCR+Image} the structured text block and the document image are provided together; for \emph{Image-Only} the OCR output is withheld and only the image is provided.


\textbf{Document Classification.} The document type is predicted by a VLM, invoked as a separate prior call using both the OCR text and the document image. The predicted class conditions downstream extraction by determining which entity schema to apply; for example, the same field name (e.g., \emph{date}) can refer to semantically different entities across document types in a multi-class packet. Because \textsc{ConfBench} contains a single document class (FCC invoice), this step always returns the correct class and classification accuracy does not vary across experimental conditions. We include it to faithfully replicate the production IDP architecture, where classification errors in multi-class packets would propagate to extraction by applying the wrong schema.



\textbf{Entity Extraction and Confidence Estimation.}
We extract entity values according to a predefined schema of target types: six document-level fields (\texttt{Agency}, \texttt{Advertiser}, \texttt{GrossTotal}, \texttt{PaymentTerms}, \texttt{AgencyCommission}, \texttt{NetAmountDue}) and a variable-length \texttt{LineItems} array, each entry containing five sub-fields (\texttt{StartDate}, \texttt{EndDate}, \texttt{Days}, \texttt{Description}, \texttt{Rate}).

How extraction and confidence estimation are coupled depends on the strategy. For \textbf{verbalized Approach} (1S-TopK method), both operations are performed jointly in a single LLM call: the model emits its top-four candidate values with associated probabilities for every entity simultaneously, and the highest-probability candidate is selected as the extraction output. For \textbf{1S-Logprob}, extraction is performed in a single greedy-decoding pass that produces a structured JSON answer; confidence scores are then derived post-hoc from the token-level log-probabilities of the tokens spanning each entity value, without any additional model call (see Appendix~\ref{sec:appendix_logprob_details} for aggregation details).


\paragraph{Evaluation.} We score extracted values using \textsc{Stickler},\footnote{\url{https://github.com/awslabs/stickler/tree/local/conf-plus-date}} an open-source structured-extraction evaluation library that requires field-level ground truth annotations. Each field uses a comparator suited to its type: string fields by character-level similarity, numeric fields by proximity within a 1\%tolerance, and date fields by semantic calendar equivalence so that surface differences in formatting count as correct. \texttt{LineItems} are matched to predictions by maximising total similarity across all line items before per-field scoring.

Each entity receives a continuous similarity score in $[0, 1]$ reflecting how closely the predicted value matches ground truth. These scores are averaged across all entities to produce WOA, our headline extraction accuracy metric. For calibration, each score is converted to a binary correctness label using a per-field acceptance threshold, and paired with the entity's confidence score. \textsc{Stickler} computes all reported metrics directly from these pairs: extraction quality (WOA, F1, precision, recall) and confidence calibration (AUROC, ECE, Brier Score, and ECARB at multiple review budgets).

\subsection{Confidence Estimation and Inference Details}
\label{sec:appendix_logprob_details}



\textbf{Verbalized 1S-TopK protocol.} Following \citet{tian2023just}, we prompt the model to produce its top-four candidate values, each with a continuous probability in $[0, 1]$, for every target entity in a single JSON call. The candidate with the highest probability is selected as the extraction output; its associated probability is retained as the confidence score and later used to assess calibration. The rationale is that forcing the model to consider alternatives reduces overconfidence compared to a single top-1 prediction; we omit chain-of-thought reasoning, as \citet{tian2023just} found it does not improve calibration further. Because extraction and confidence estimation are coupled in the same call, entity-level accuracy varies across models, unlike logprob methods where extraction is always decoupled from confidence scoring.

\revision{
\paragraph{Coupling of extraction and confidence estimation.} The two approaches differ in whether the model sees the confidence task while it is extracting, which is why the same model can report a slightly different extraction accuracy under each. Under 1S-Logprob the confidence score is derived post-hoc from the token log-probabilities of a standard greedy-decoding pass. The prompt asks only for the JSON answer, the decoding trajectory is exactly the one the model would follow if confidence were never requested, and the scores are read off afterwards from the stored logits. Extraction accuracy therefore reflects pure task performance. Under verbalized 1S-TopK, extraction and confidence estimation happen jointly in one call: the model must enumerate four candidate values per entity and attach a probability to each, and the top-ranked candidate doubles as the extraction output. The coupling is intentional and follows \citet{tian2023just}: requiring the model to weigh alternatives against one another is what yields better-calibrated verbalized confidence for capable models, and splitting the call into two would discard that mechanism. The cost is that the confidence task now sits inside the extraction context, so it can shift what gets extracted---either helping, by making the model deliberate over near-miss candidates, or hurting, by consuming output budget that would otherwise go to the answer itself. Only two models in our study admit both approaches, and Table~\ref{tab:results-ocr-image} shows small differences in opposite directions for those two, so this evidence cannot settle whether joint estimation helps or hurts extraction; the effect, if any, appears to be model-specific. Isolating it requires a fully decoupled two-stage verbalized design---extract first, then score the confidence of the produced output in a separate call, or with a separate model---which holds the extraction pass fixed across confidence strategies in the way logprob already does. That is a natural next step, and it also raises the question of whether a separate scoring model is better calibrated on another model's output than the extractor is on its own.
}

\revision{
\paragraph{Model coverage for logprob confidence.} Logprob confidence requires access to token-level log-probabilities over the vocabulary at each generation step. Commercial inference APIs do not expose this for the vision-language models we evaluate, so the approach is available only for models we self-host. This is why 1S-Logprob is reported for Qwen~3.6-27B and Gemma~3-12B but not for Qwen~3~VL-235B: its parameter count places self-hosted inference beyond the accelerator memory available to us, and serving it for the full evaluation was computationally prohibitive under our infrastructure constraints. Qwen~3~VL-235B is therefore evaluated only under verbalized 1S-TopK, which needs nothing beyond the text of the response and so runs unchanged against a hosted endpoint. The gap is an infrastructure limitation, not a modeling result: nothing about the model precludes logprob confidence, and the comparison would be informative given that it is the largest open-weight model in the study.
}


\paragraph{Aggregation methods for 1S-Logprob.}
Let $\{\ell_1, \ldots, \ell_K\}$ be the token log-probabilities of the $K$ generated tokens covering an entity value. Separately, let $p_1 \geq p_2$ denote the top-2 probabilities over the \emph{full vocabulary} at the first token position, that is, the probability of the chosen first token and the probability of the next-best alternative at that same position, not the log-probs of the first and second tokens in the sequence. We compare three ways of summarizing these into a single confidence score.
\begin{itemize}
    \item \textbf{Mean-token} $\exp\!\left(\tfrac{1}{K}\sum_k \ell_k\right)$ is the geometric mean probability across all tokens in the value, capturing overall generation confidence.
    \item \textbf{First-token} $\exp(\ell_1)$ uses only the probability of the first token, reflecting how decisively the model committed to this value at the point of first generation.
    \item \textbf{Margin} $\tfrac{p_1 - p_2}{p_1 + p_2}$ is the normalized gap between the top-2 vocabulary candidates at the first token position, measuring how unambiguously the model chose its answer over the next best alternative.
\end{itemize}

\paragraph{Token budgets.}
For verbalized 1S-TopK, output token limits follow the model provider. Anthropic models (Claude Opus, Sonnet, Haiku) are capped at 40K tokens, reflecting Bedrock's higher output limit for those models. All open-weight models are capped at 8K tokens, matching the Bedrock output limit for those providers; self-hosted models (Qwen~3.6-27B, Gemma~3-12B) use the same 8K cap via \texttt{max\_new\_tokens} for consistency. Because the model emits four candidates per entity across all fields in a single response, entity counts can vary across models due to differences in output verbosity under these caps. For 1S-Logprob, self-hosted models use \texttt{max\_new\_tokens}~$= 8192$, which is sufficient because the prompt requests a single compact JSON answer rather than multiple candidates.

\revision{
\paragraph{What the budgets do and do not control.} Each model's output cap is held constant across both confidence approaches, so \emph{within}-model comparisons of verbalized against logprob confidence are fully controlled: the same model sees the same documents under the same budget, and the only thing that changes is how confidence is obtained. \emph{Across} models the comparison is only partially controlled, because the caps themselves differ (40K for Anthropic models, 8K for open-weight models). The lower open-weight cap is a deliberate design choice and not a convenience. Those models degenerate into repetition on certain fields, most often the nested \texttt{LineItems} sub-fields, and raising the ceiling extends the repeated span instead of yielding additional valid entities; a larger budget would therefore add cost and latency while worsening output quality rather than improving coverage. One consequence is that the number of evaluated entities is not identical across models. The dominant source of that variation is again \texttt{LineItems}, a nested variable-length list whose length differs per document: models differ in how many items they enumerate before repeating or truncating, so a model that walks the whole table cleanly contributes more entities than one that stalls partway. Every confidence metric we report---AUROC, ECE, Brier, and ECARB---is computed per entity and then aggregated, so the differing evaluation-set sizes do not bias the calibration scores themselves, and a model gains nothing by extracting fewer entities. The sizes do vary, however, so per-model estimates rest on different numbers of entities, and we accordingly avoid reading small cross-model gaps as strict orderings.
}


\textbf{Image resize.}
For 1S-TopK, we resize document images to $1200 \times 1000$ pixels, preserving aspect ratio, to give verbalized models sufficient visual resolution for fine-grained text recognition. For 1S-Logprob we set \texttt{max\_pixels}~$= 800 \times 800 = 640{,}000$ in the Qwen processor. This is a total pixel budget: the processor dynamically rescales the image so that width~$\times$~height does not exceed this limit while preserving the original aspect ratio, rather than forcing a fixed output shape. This reduces memory footprint during greedy decoding without materially affecting extraction quality on single-page invoice documents.

\textbf{Generation hyperparameters for 1S-Logprob.} We run greedy decoding with temperature $T{=}0$ and \texttt{top\_p}\,$=$\,0.1 on Qwen~3.6-27B and Gemma~3-12B, capturing the top-10 logprobs at each generation step from the raw pre-temperature logits via \texttt{torch.log\_softmax}. Chain-of-thought is disabled (\texttt{enable\_thinking=False} for Qwen~3.6; Gemma~3 does not have a CoT toggle) so logprobs cover only the JSON answer.

\textbf{Value-token alignment.} Given the generated token sequence and its decoded text, we (1)~locate the field key \texttt{"FieldName"} in the reconstructed string, (2)~find the value substring after the key, (3)~map each character offset to its source token index via a precomputed character-to-token map, and (4)~deduplicate and order the resulting indices. For repeated keys inside the \texttt{LineItems} array, we anchor on the $n$-th occurrence of \texttt{"LineItemStartDate"} to disambiguate the target line item, then search forward for the requested sub-field.

\textbf{Prompt adaptation across providers.} Prompts are reformatted to match each provider's expected schema (system / user roles, JSON-mode flags, image-token placement) while preserving the underlying instructions, output schema, and field descriptions, so that differences across models reflect model capability rather than prompt engineering.

\subsection{Interpreting ECARB}
\label{sec:appendix_ecarb}

\revision{
\paragraph{What the metric measures.} ECARB@$b$ answers a single operational question: for a fixed human review budget, how much more effective is confidence-guided review than random sampling at surfacing errors? Given $n$ predictions sorted by ascending confidence, the $k = \min(n, \max(1, \lfloor bn \rfloor))$ least-confident predictions are routed to human review. A reviewer who instead sampled $k$ predictions at random would catch $(k/n) \cdot E$ of the $E$ total errors in expectation. ECARB@$b$ is the ratio of the errors actually caught in the low-confidence slice to that random-sampling expectation, as defined in \S\ref{sec:foundation_models}. It is a lift, not an error rate, so it does not depend on how many errors the model makes overall, only on how well confidence concentrates those errors at the bottom of the ranking.
}

\revision{
\paragraph{Reading the values.} The reference point is $1$, not $0$:
\begin{itemize}
  \item $\text{ECARB}@b = 1$: confidence-guided review catches exactly as many errors as random review. The scores carry no targeting benefit, and sorting by confidence is wasted effort.
  \item $\text{ECARB}@b > 1$: more errors are surfaced per reviewed item than random sampling would surface, so the scores usefully direct HITL effort. Higher is better.
  \item $\text{ECARB}@b < 1$: fewer errors than random. This is the failure case, indicating scores that are uninformative or negatively correlated with correctness at the reviewed end of the ranking. No configuration in Table~\ref{tab:results-ocr-image} falls below $1$; the lowest value observed is $1.30$.
\end{itemize}
The metric is bounded above by $\min(1/b, n/E)$, the point at which the reviewed slice contains every error (or consists entirely of errors), so the achievable ceiling tightens as the review budget shrinks.
}

\revision{
\paragraph{Worked example.} Consider a run with $n = 100$ predictions of which $E = 20$ are errors, evaluated at a review budget of $b = 0.30$, so that $k = 30$ predictions are reviewed. A random reviewer examining 30 of the 100 predictions catches
\[
(k/n) \cdot E \;=\; 0.30 \times 20 \;=\; 6
\]
errors in expectation. Suppose the 30 lowest-confidence predictions instead contain 15 of the 20 errors. Then
\[
\text{ECARB}@0.30 \;=\; \frac{15}{6} \;=\; 2.5 ,
\]
that is, confidence-guided review surfaces $2.5\times$ as many errors as random review for the same reviewer effort. Here the ceiling is $\min(1/0.30, 100/20) = 3.33$, attained only if all 20 errors were ranked among the 30 least-confident predictions. Applied to Table~\ref{tab:results-ocr-image}, Opus~4.6's $\text{ECARB} = 2.43$ means that reviewing the least-confident 30\% of its predictions catches $2.43$ times as many errors as reviewing a random 30\% of them.
}

\revision{
\paragraph{Choice of review budget.} We report $b = 0.30$ throughout as a single representative operating point, not as a recommended review rate; the appropriate budget is set by the cost of review and the error tolerance of the deployment, and practitioners should evaluate ECARB at their own $b$. Because ECARB depends only on the ordering that confidence induces over predictions and not on the absolute confidence values, any strictly monotone rescaling of the scores leaves it unchanged, and model rankings by ECARB are largely stable as the budget varies. What changes with $b$ is the magnitude of the lift and its ceiling, not which models rank above which.
}

\revision{
\paragraph{Reading Table~\ref{tab:results-ocr-image}.} The six reported columns answer different questions and are not meant to be weighed equally. ECARB speaks most directly to the deployment question, ``which configuration catches the most errors for my review budget?'', and is the column to optimize when predictions are reviewed in confidence order. WOA and AUROC form the diagnostic decomposition behind it, explaining why two configurations with similar ECARB arrive there by different routes: fewer total errors to find versus sharper discrimination among the errors that exist. ECE matters only when routing uses an absolute confidence threshold rather than a ranked queue, since it measures how far stated confidences sit from observed accuracies. Brier is partly redundant once AUROC and ECE are reported separately, as it mixes discrimination and calibration into one number; we include it as a proper-scoring-rule sanity check that is free of the binning choices ECE depends on.
}

\revision{
\paragraph{Why AUROC and ECE are both needed.} ECARB is a ranking-driven quantity and therefore tracks AUROC, which measures discriminative ability: whether correct predictions receive higher confidence than incorrect ones. AUROC says nothing about whether the confidence values themselves are close to the corresponding empirical accuracies. ECE measures exactly that gap and says nothing about discrimination. The two can diverge in either direction. A model that assigns confidence $0.99$ to every correct prediction and $0.01$ to every incorrect one achieves perfect AUROC but poor ECE, because its stated confidences are far from the observed accuracies in each bin, even though the ranking is flawless. Conversely, a model that assigns $0.70$ to every prediction in a dataset where $70\%$ of predictions are correct achieves near-zero ECE while its AUROC sits near $0.5$, because a constant score cannot separate correct from incorrect predictions at all. This is why we report both, and why the two orderings in Table~\ref{tab:results-ocr-image} do not coincide: high AUROC with high ECE identifies scores that rank well but whose absolute values should not be used as decision thresholds without correction, and the reliability diagrams in \S\ref{sec:appendix_reliability} show where each model's stated confidence departs from observed accuracy.
}

\subsection{Scope of Cross-Model Observations and Deployment Guidance}
\label{sec:appendix_analysis_notes}

\revision{
This subsection expands the points that \S\ref{sec:results} states in compressed form: how far our cross-model observations generalize, what post-hoc calibration can and cannot fix, how to read the gap between WOA and F1, and how to weigh extraction accuracy against confidence quality when selecting a configuration.
}

\revision{
\paragraph{Architecture family.} Within our model pool, the two MoE models (Kimi~K2.5 and Qwen~3~VL-235B) trail dense models of comparable active parameter count on AUROC. Two models cannot support an architecture-level claim, so we report this as a property of the specific systems evaluated and do not attribute the gap to sparse activation. These models also differ from the dense ones in training data, instruction tuning, and release generation, any of which could account for the difference. Establishing whether sparse activation itself affects confidence quality would require a controlled comparison of dense and sparse variants trained on matched data, which no currently available model family provides at the scales we evaluate.
}

\revision{
\paragraph{Capability and the preferred confidence family.} Only Qwen~3.6-27B and Gemma~3-12B expose token-level probabilities, so the comparison of verbalized against logprob confidence rests on two models, and they disagree on which family wins: verbalized leads for Qwen~3.6-27B (0.72 against 0.62 AUROC) while logprob first-token leads for Gemma~3-12B (0.64 against 0.58). A plausible reading is that the weaker model cannot self-assess reliably while still producing informative logit distributions, but a two-point pattern is not a general relationship between capability and the preferred confidence family. We therefore treat the choice as model-dependent and something to settle per model on held-out data rather than by rule. Logprob carries one practical advantage independent of which family scores better: it needs no prompt engineering, and it can be read off a decoding pass the pipeline is already running.
}

\revision{
\paragraph{What post-hoc calibration can and cannot fix.} The reliability diagrams bound what post-hoc calibration can buy. Rescaling confidence values improves ECE, which matters when routing uses an absolute threshold, for instance auto-accepting every prediction above 0.9. Rescaling is monotone, however, so it does not reorder predictions, which leaves AUROC and ECARB essentially unchanged. Calibration is therefore situational rather than a blanket requirement: a deployment that reviews a fixed fraction of its queue in confidence order gains nothing from it, while one that routes on an absolute threshold needs it. For the overconfident configurations, a per-bin correction can be read straight off the diagrams in Appendix~\ref{sec:appendix_reliability} without fitting a calibrator. We do not report calibrated results, since doing so would require a held-out calibration split that our corpus size does not comfortably support.
}

\revision{
\paragraph{Imperfect extraction strengthens the case for confidence routing.} The F1 column in Table~\ref{tab:results-ocr-image} sits well below WOA because the two score the same predictions on different scales rather than because the task is failing (\S\ref{sec:foundation_models}): WOA credits partial matches continuously, while F1 first thresholds each field into accept or reject. Read even at its strictest, a moderate error rate is the regime in which confidence routing is most useful. Were extraction near-perfect, a reviewer would have little left to find and the ranking would hardly matter, whereas a substantial pool of errors is what confidence-ordered review is designed to prioritize, and ECARB shows that confidence does surface it, up to 2.43$\times$ random at a 30\% budget. Organizations deploy VLMs at whatever capability is currently available, so the operational question is not whether errors remain but how cheaply they can be located.
}

\revision{
\paragraph{Selecting a configuration.} The guidance in \S\ref{sec:results-metrics} is scoped to HITL workflows in which predictions are reviewed in ascending confidence order. Extraction accuracy and confidence quality answer complementary questions and neither subsumes the other. WOA fixes how many errors a batch contains, that is, the pool to be found; AUROC and ECARB fix what fraction of that pool confidence-guided review surfaces per unit of reviewer effort. Where one model has substantially lower WOA than another and their AUROCs are comparable, the accuracy gap dominates and the more accurate model should be preferred. Conversely, a model with higher accuracy but poorly ranked confidence can leave more errors undiscovered at a fixed budget than a slightly less accurate model whose confidence ranks errors well. Our results show how far the two can diverge: Sonnet~4.5 attains the highest extraction accuracy at 0.77 WOA yet Opus~4.6 leads on every confidence metric, and Kimi~K2.5 reaches competitive accuracy at 0.76 WOA while lagging in AUROC at 0.67. We therefore suggest shortlisting on WOA and AUROC together, adding post-hoc calibration only where threshold-based routing is required, and reading ECARB at the review budget the deployment can actually staff rather than at the 30\% we report.
}

\subsection{Reliability Diagrams}
\label{sec:appendix_reliability}

\begin{figure*}[p]
  \centering
  \includegraphics[width=\textwidth,height=0.85\textheight,keepaspectratio]{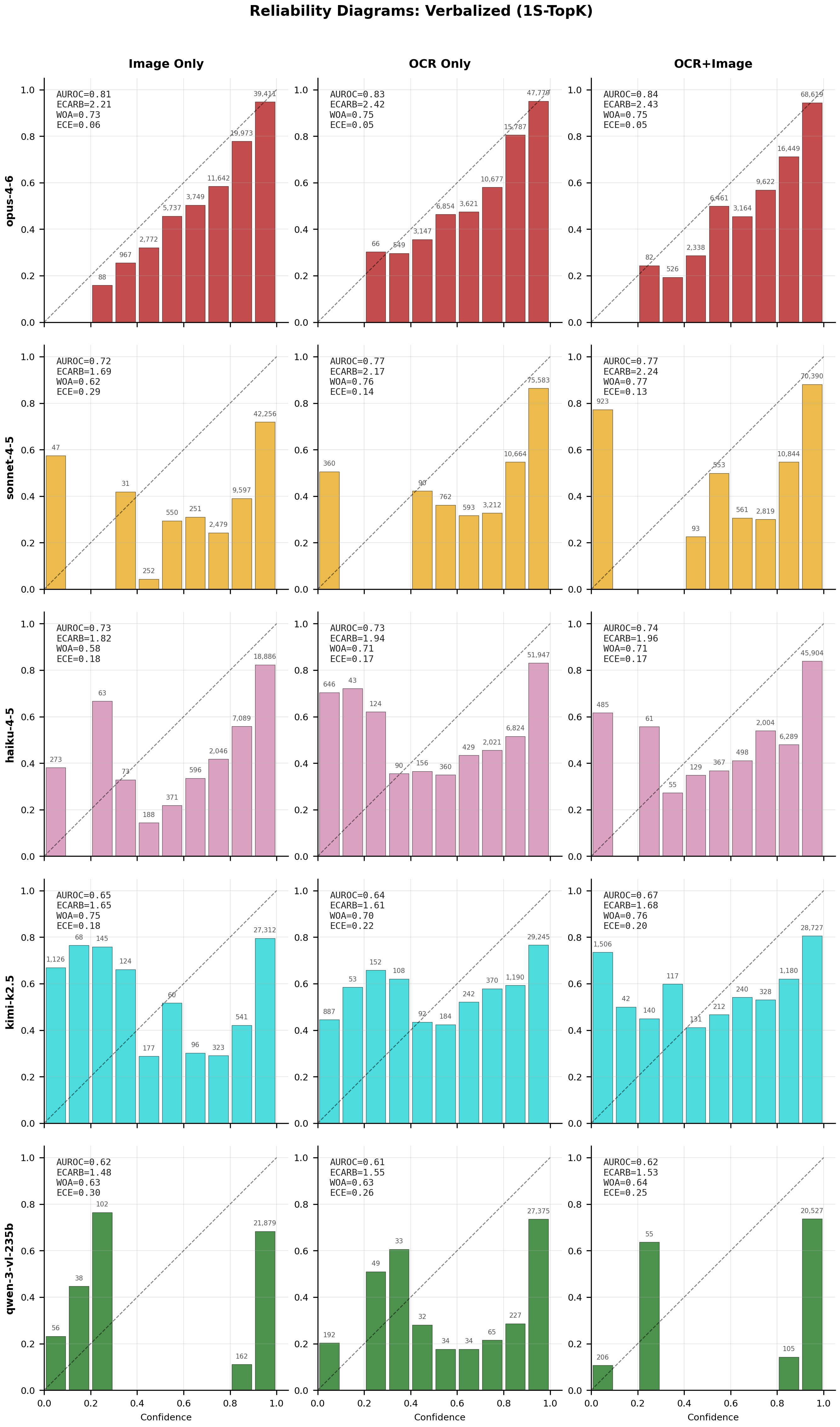}
  \caption{Reliability diagrams for all verbalized (1S-TopK) configurations. Rows: models ordered by AUROC (top = best). Columns: Image Only (left), OCR Only (center), OCR+Image (right). Opus~4.6 closely tracks the diagonal across all modalities, while weaker models show progressively larger overconfident deviations. The dominant high-confidence bin (0.9--1.0) contains the majority of predictions for all models.}
  \label{fig:reliability-verbalized}
\end{figure*}

\begin{figure*}[p]
  \centering
  \includegraphics[width=\textwidth,height=0.85\textheight,keepaspectratio]{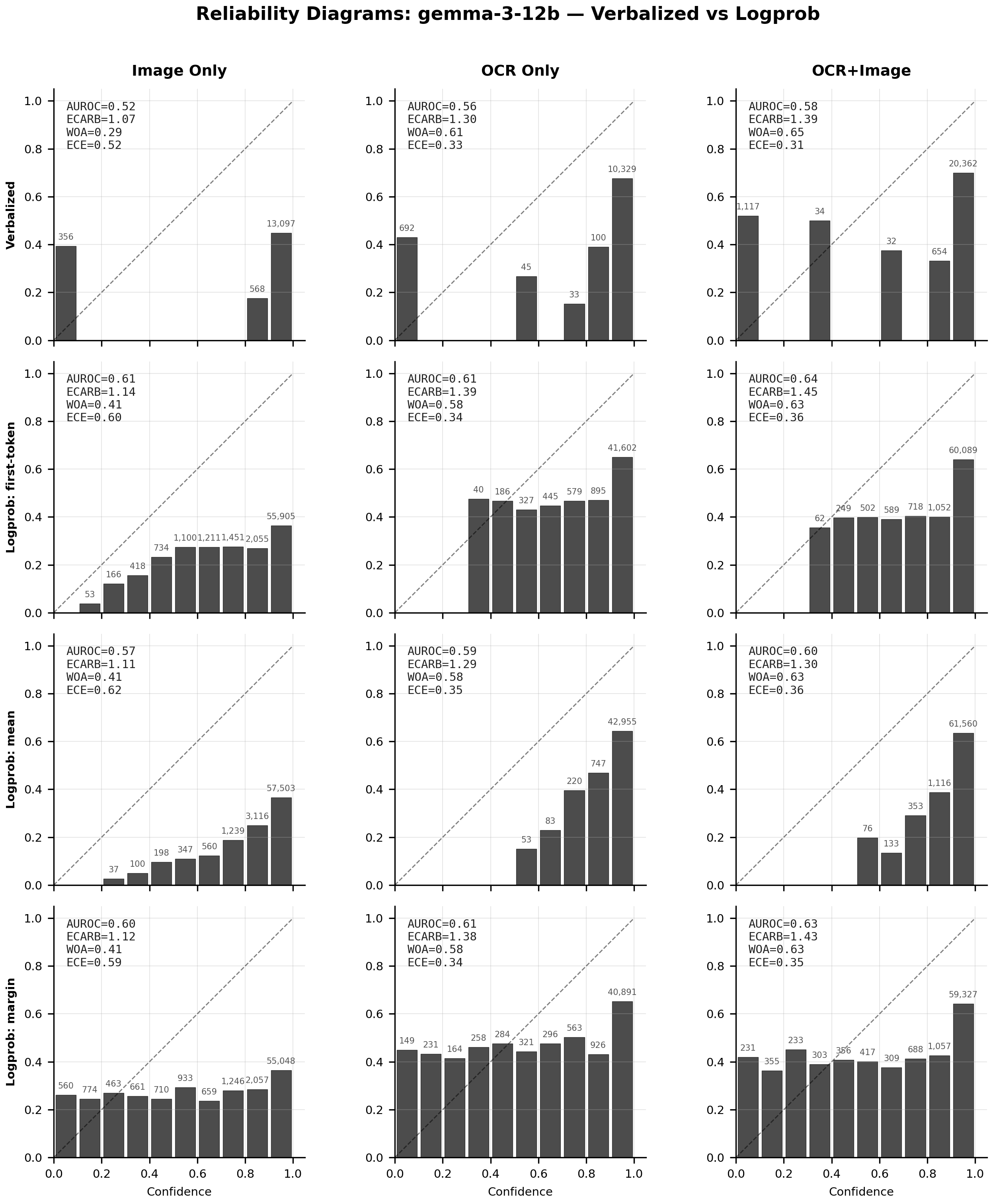}
  \caption{Reliability diagrams for Gemma~3-12B: verbalized vs logprob strategies. Rows (top to bottom): verbalized, first-token, mean, margin. Columns: Image Only (left), OCR Only (center), OCR+Image (right). First-token achieves the best AUROC. Both verbalized and first-token methods concentrate predictions in high-confidence bins. Mean-token tracks the y=x diagonal more faithfully. Margin shows the least alignment between the confidence measure and the corresponding accuracy.}
  \label{fig:reliability-logprob-gemma}
\end{figure*}

\begin{figure*}[p]
  \centering
  \includegraphics[width=\textwidth,height=0.85\textheight,keepaspectratio]{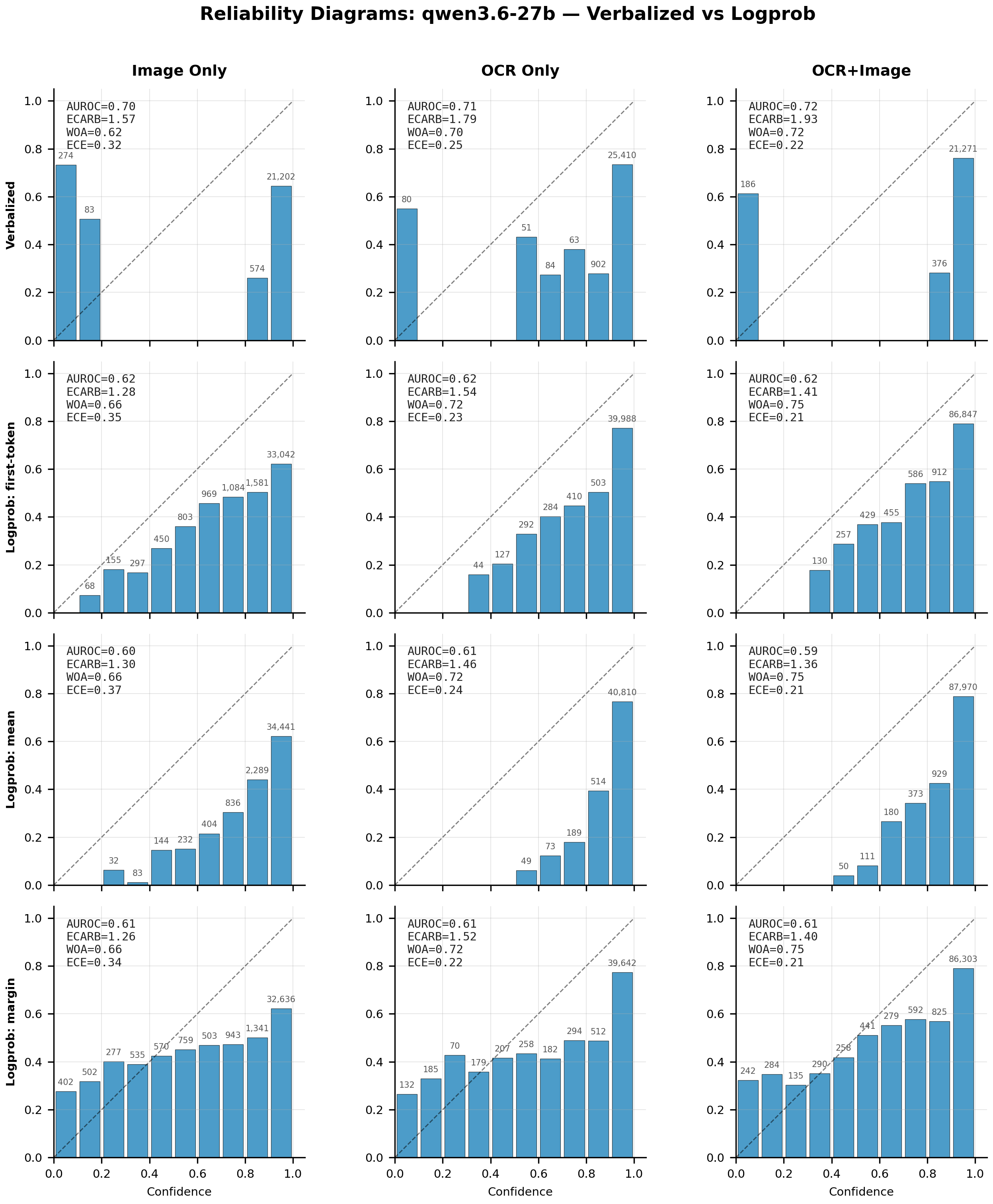}
  \caption{Reliability diagrams for Qwen~3.6-27B: verbalized vs logprob strategies. Rows (top to bottom): verbalized, first-token, mean, margin. Columns: Image Only (left), OCR Only (center), OCR+Image (right). Verbalized achieves the highest AUROC with extreme bimodal distribution. Among logprob methods, first-token and mean show similar ECE but mean tracks the diagonal more closely than the other two methods.}
  \label{fig:reliability-logprob-qwen}
\end{figure*}

Figures~\ref{fig:reliability-verbalized}, \ref{fig:reliability-logprob-gemma}, and~\ref{fig:reliability-logprob-qwen} present reliability diagrams for all evaluated configurations. While AUROC captures discriminative performance (the ability to separate correct from incorrect predictions), reliability diagrams reveal how well confidence scores are aligned with actual accuracy. The y=x diagonal represents ideal calibration; points below it indicate overconfidence, while points above indicate underconfidence. Notably, a model can have high AUROC (good discrimination) but poor calibration, or vice versa. Each panel plots accuracy (y-axis) against confidence (x-axis). Bins with fewer than 30 samples are omitted. Numbers above bars indicate sample counts per bin.

\textbf{Key observations.}
(1)~\emph{Modality effect on calibration:} Image-only consistently yields the worst ECE across all models, while OCR+Image produces similar or better calibration than OCR-only, confirming that textual OCR signals improve not only discriminative power but also absolute calibration alignment.
(2)~\emph{Verbalized overconfidence:} For the two models studied in direct strategy comparison (Gemma~3-12B and Qwen~3.6-27B), verbalized confidence concentrates the vast majority of predictions in the highest confidence bin (0.9--1.0). ECE is therefore dominated by the accuracy of this single bin.
(3)~\emph{Mean-token follows y=x better than first-token:} Despite first-token achieving higher AUROC, mean-token logprob produces bars that track the diagonal more closely across mid-range bins. The similar ECE values arise because first-token's dominant high-confidence bin happens to be well-calibrated, masking its deviations elsewhere.

\end{document}